\documentclass[fleqn,10pt]{wlscirep}
\usepackage[utf8]{inputenc}
\usepackage[T1]{fontenc}
\usepackage{placeins} 
\usepackage{amsmath}
\usepackage{bbm}
\usepackage{hyperref}
\usepackage[textsize = tiny,backgroundcolor=blue!10!white]{todonotes}

\title{The Inner Sentiments of a Thought}
\author[1,2,*]{Chris Gagne}
\author[1,3]{Peter Dayan}
\affil[1]{MPI for Biological Cybernetics, Tübingen, Germany}
\affil[2]{Research Division, Hume AI, New York, USA}
\affil[3]{University of Tübingen, Tübingen, Germany}
\affil[*]{christopher.gagne@tuebingen.mpg.de}

\DeclareGraphicsExtensions{.png,.pdf}
\newcommand{\cut}[1]{}

\begin{abstract}

Transformer-based large-scale language models (LLMs) are able to generate highly realistic text. They are duly able to express, and at least implicitly represent, a wide range of sentiments and color, from the obvious, such as valence and arousal to the subtle, such as determination and admiration. We provide a first exploration of these representations and how they can be used for understanding the inner sentimental workings of single sentences. We train predictors of the quantiles of the distributions of final sentiments of sentences from the hidden representations of an LLM applied to prefixes of increasing lengths. After showing that predictors of distributions of valence, determination, admiration, anxiety and annoyance are well calibrated, we provide examples of using these predictors for analyzing sentences, illustrating, for instance, how even ordinary conjunctions (e.g., “but”) can dramatically alter the emotional trajectory of an utterance. We then show how to exploit the distributional predictions to generate sentences with sentiments in the tails of distributions. We discuss the implications of our results for the inner workings of thoughts, for instance for psychiatric dysfunction.

\end{abstract}
\begin{document}

\flushbottom
\maketitle

\section*{Introduction}

How do sentences do affective work? That is, how do sentences, with their complex syntax and semantics, toy with our expectations in order
to pack an emotional punch? How could we generate new sentences that are
finely pitched at a particular strength of sentiment?  Equally, what
does someone's choice of sentences reveal about their psychological
state? If even the idiosyncratic choice of innocuous function words
(e.g., pronouns) can be used to predict affective states,
from the momentary to the more long lasting, such as depression
\cite{chung2007psychological}, how much more might be encoded in the
detailed affective topography of someone's utterances?

Until recently, we have lacked access to the inner semantic and syntactic content of sentences required to perform fine-grained, automatic analyses of their emotional dynamics, and thus answer these
questions. However, the advent of large-scale pretrained language models (LLMs)\cite{vaswani2017attention, devlin2018bert, radford2019language} has dramatically altered the scene. The hidden activities in these models as they `read' text have been used as representations to predict, with incredible accuracy, a wide range of linguistic features, from the parts-of-speech of individual words to logical entailment of adjacent utterances. They have been used to perform various tasks like question-answering, text-retrieval, sentiment analysis, and document clustering  \cite{muennighoff2022mteb, srivastava2022beyond}. And of course, they have been used to generate uncannily human-like sentences, paragraphs, (bad) limericks, and even articles. This has led to a flurry of papers trying to understand how LLMs work, the linguistic information encoded their hidden activities (or attention weights) \cite{clark2019does, tenney2019bert}, and whether they `understand' language \cite{rogers2021primer, pavlick2022semantic, li2022systematicity} (or even reason)\cite{binz2022using, dasgupta2022language} like we do. Comparatively less emphasis has been placed on the other direction -- using LLMs to understand how text, on a micro-level, can be used to convey information like emotional sentiment.

Here we take initial steps in this direction. We use LLMs to provide an embedding space for the state within an utterance, and train models to predict from this the quantiles of the distribution of the \emph{end} ratings of emotionally relevant sentiments. The evolution of the resulting quantile predictions in new sentences tells us how the sentence is moving in high and low-dimensional emotion space -- showing clearly, for instance, how the conjunction “but” can suddenly reverse the sentiment or cue the imminent rise of more subtle emotional tones. Equipped with predictive distributions of emotion, we then demonstrate how to generate text that targets particular affective quantiles, in a method that relates to long-run risk-sensitive choice. Although many recent methods (finetuning, prompting, etc.) have been developed to alter the writing style of LLMs \cite{krause2020gedi, holtzman2018learning, yang2021fudge, snell2022context, snell2022offline, guo2021text, leblond2021machine, stiennon2020learning, ouyang2022training}, our method uniquely provides fine-grained control over particular quantiles for a wide range of emotional sentiments. 


\section*{Results}

\subsection*{Predicting distributions of sentiment}

Our testbed is the abundant, and often emotionally-charged, discourses of people on Reddit \cite{cohan2018smhd}. We first extract the hidden states of a large-scale language model (we chose the comparatively venerable GPT-2\cite{radford2019language} for availability and computational practicality) applied to 2 million sentences. We then train a set of smaller models to predict the quantiles of the distributions of the end ratings for these sentences along several different emotionally-relevant dimensions: positive and negative valence, determination, admiration, annoyance, and anxiety. A separate model was trained for each of these dimensions. We chose these dimensions to highlight the applicability of our method to both low-dimensional (i.e., valence and arousal) and high-dimensional models of emotion. The end ratings were obtained automatically using other language models \cite{barbieri2020tweeteval, demszky2020goemotions}, which had been fine-tuned to predict human emotional judgements of short whole utterances.

\begin{figure}[ht]
\centering
\includegraphics[width=\linewidth]{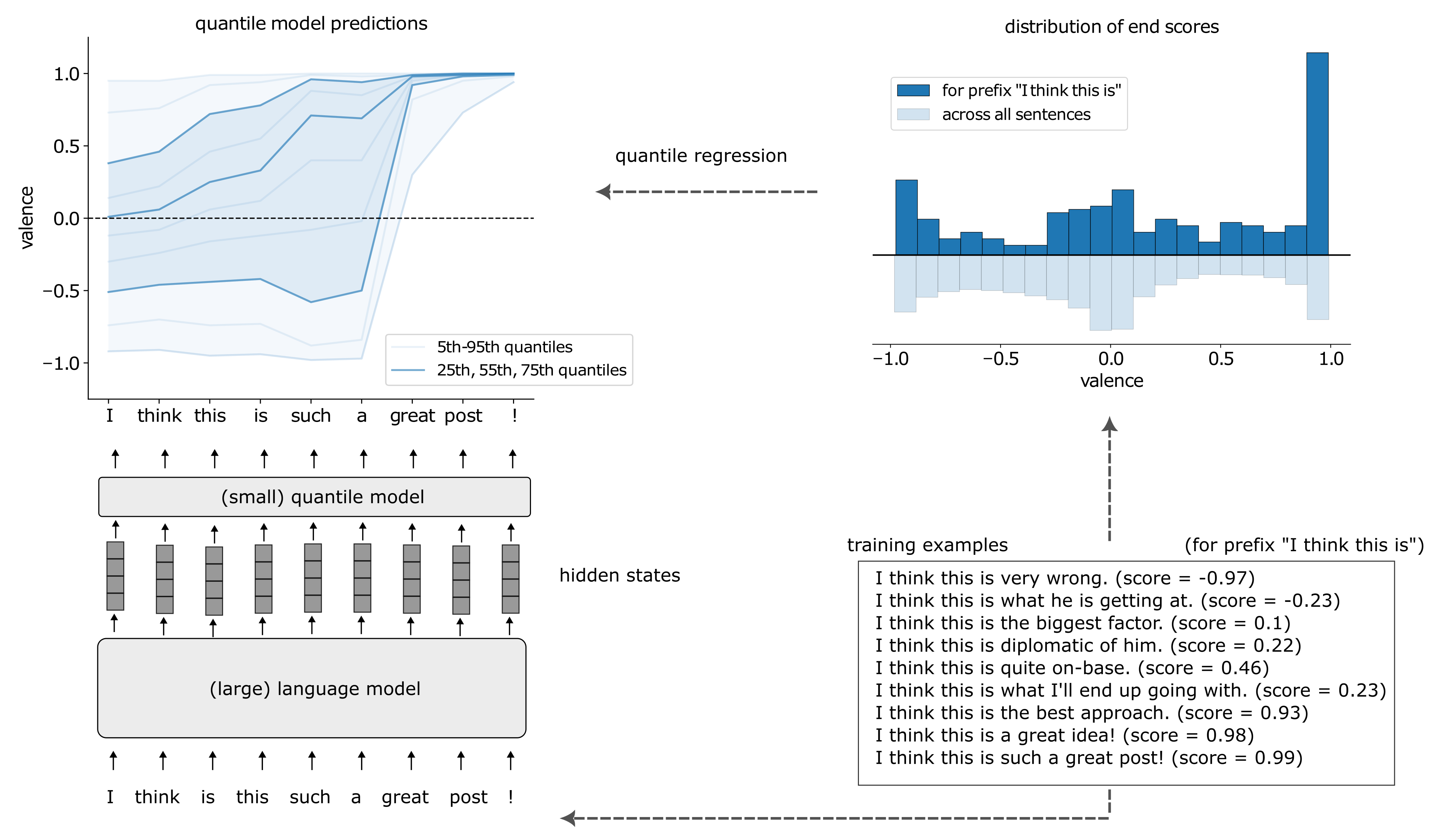}
\caption{\textbf{Training a model to predict distributions of valence and emotion throughout a sentence.} Tokenized sentences from Reddit are fed into a large language model (lower left), and the hidden states at each token are used to predict the quantiles (upper left) of the distribution of end scores. End scores are provided by a sentiment model (or emotion classification model) applied to the full sentence. Quantile regression is used to train the model on the prefixes of increasing lengths (e.g., “I”, “I think”, etc) using the empirical distribution of end scores for sentences starting with those prefixes, and therefore the model is able generate a trajectory of predicted quantiles across successive tokens. The empirical distribution of valence scores and examples for the prefix “I think this is” are shown on the right. The marginal distribution (i.e., the scores across all sentences) is shown inverted and underneath. Empirical distributions and predictions for the other emotions are shown in Figure ~\ref{fig:ngrams}.}
\label{fig:training}
\end{figure}

As depicted in Figure~\ref{fig:training}, the quantile models are provided with the hidden states from GPT-2 for each token, and are trained using Monte Carlo methods and quantile regression\cite{Koenker2005, dabney2018distributional, bellemare2023distributional} to predict the distribution of \emph{end} scores. Scores are only provided at the end of the sentence, and therefore the quantile model must predict how the sentence might end for each prefix (i.e., at each token position). For some common prefixes, such as “I think this is …”, the model encounters an empirical distribution of different possible end scores in the dataset (Figure~\ref{fig:training}; top right) and can learn to approximate it. For other prefixes, the model may only see a single example and its score, and therefore must learn to generalize across utterances with similar emotional potentials. The model generates a trajectory of predicted quantiles across the successive tokens (Figure~\ref{fig:training}; top left). Observe how the width of the predicted valence distribution varies throughout the sentence, reaching a maximal width at the token “such”, and then collapses around a single (correct) predicted score by the end as the final sentiment is certain. As seen in the figure, the quantile model can capture bi- and multi-modal predictions.

\subsubsection*{Model calibration}

To evaluate the accuracy of the predicted quantiles, we use 1M validation sentences that were unseen during training. We first assess global (marginal) calibration by calculating the percentages of end scores that fall below each of the predicted quantiles, aggregated across all sentences. For a perfectly calibrated model, the end scores would fall below the $\alpha$-quantile exactly $\alpha$-\% of the time, regardless of the fact that the quantile predictions correspond to different sentences and different prefixes within a sentence. In Supplemental Figure~\ref{fig:calibration}, we show that the models are highly accurate, with calibration curves barely deviating from the identity line and with a maximum absolute deviation less than 1\% for valence and between 2-7\% for the four emotions, which have much sparser empirical distributions of end scores (Supplemental Figure~\ref{fig:marginal_dists}). The models are well-calibrated even at the start of the sentence, meaning that they can accurately predict the distribution a dozen or more words ahead (since the median sentence length is 15). As an alternative to Monte Carlo methods, we also investigated temporal difference (TD) learning for training the quantile models\cite{bellemare2023distributional}; however, the resulting models were more poorly calibrated, with a maximum absolute deviation of 9\% for valence (see Methods for more details).

\begin{figure}[ht]
\centering
\includegraphics[width=\linewidth]{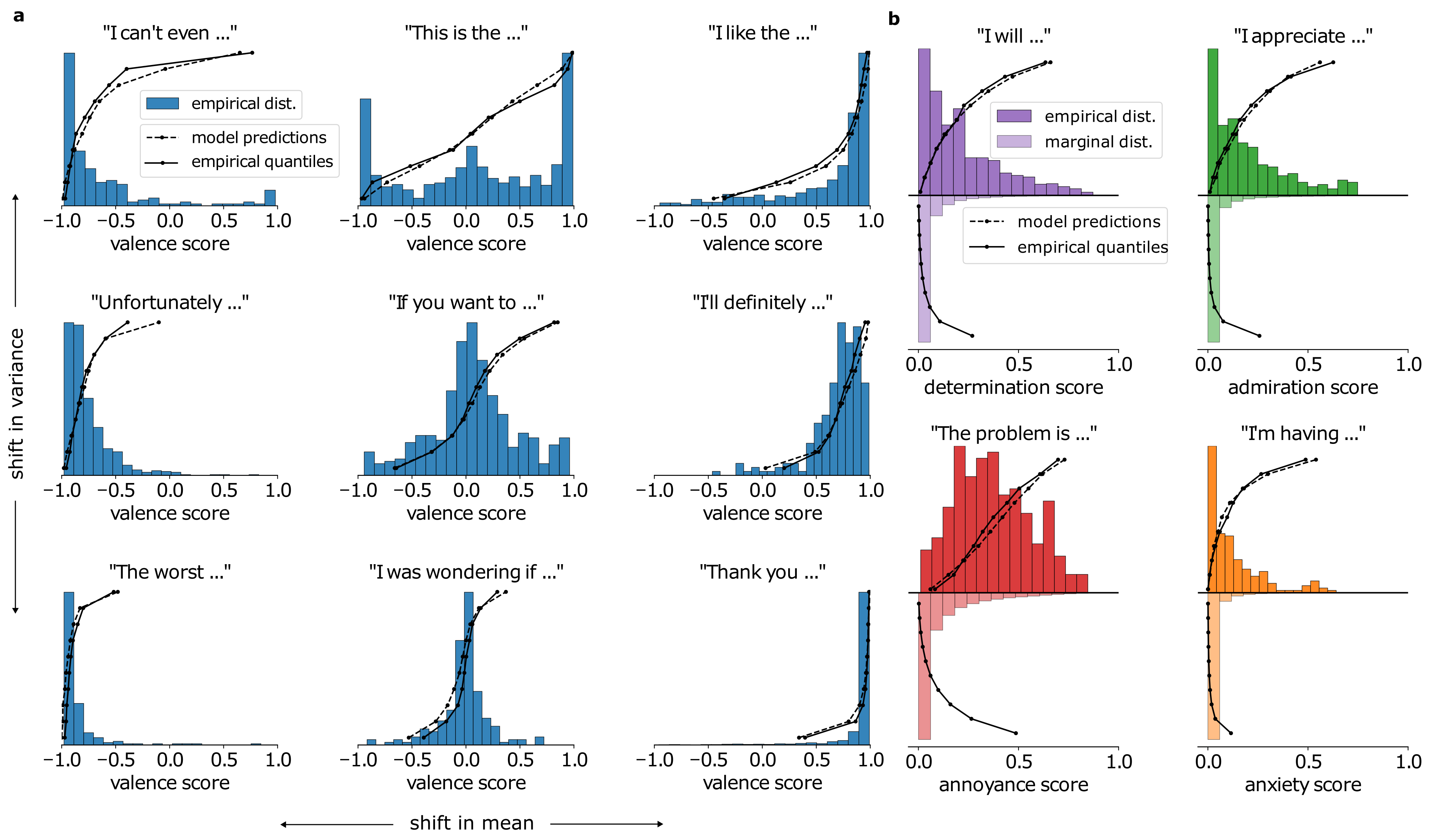}
\caption{\textbf{Model predictions match empirical distributions for short prefixes.} (a) The predicted quantiles for several example prefixes are plotted against the empirical quantiles (evaluated using\cite{barbieri2020tweeteval, demszky2020goemotions}) from the validation sentences for valence; examples are organized according to the mean and variance of the distribution to show the model's flexibility. (b) Emotion scores have much sparser and more skewed marginal distributions (shown inverted underneath) than valence; nevertheless the predicted quantiles match the empirical distributions well. The error (average maximum deviation) between the empirical and estimated quantiles for these examples is 0.14, and it is 0.1 for the top 1000 most often occurring prefixes, demonstrating the representativeness of these examples. 
}
\label{fig:ngrams}
\end{figure}

We also compared the models’ predicted quantiles to the empirical distributions of end scores for frequently occurring prefixes. Figure~\ref{fig:ngrams}a contains several representative examples, which are organized according to the mean (columns) and variance (rows) of the predicted distribution. The model predictions for these prefixes closely match the quantiles of the empirical distributions of end valence scores, despite these distributions differing considerably from the marginal distribution (shown in Figure~\ref{fig:training} and Supplemental Figure~\ref{fig:marginal_dists}). This demonstrates the model's ability to capture distributions of non-trivial shapes, from those that are highly skewed (e.g., “Thank you …”) to those that are more bimodal (e.g., “This is the …”). Furthermore, note that although these examples were hand-selected, their estimation error (average maximum deviation between quantile sets) does not differ to that of the top 1000 most frequently occurring prefixes.

In Figure~\ref{fig:ngrams}b, we plot examples for each emotion, showing large positive shifts in the distribution for phrases such as “I will …” (for determination), “I appreciate …” (for admiration), “The problem is …” (for annoyance), and “I'm having …” (for anxiety). Note that the marginal distributions, shown in lighter color beneath each example, are much more skewed for these four emotions than for valence -- and yet the models are still able to make accurate predictions. This means that quantiles could also be estimated for other semantic characteristics that have sparse distributions, such as toxicity or conversational topic\cite{mathewson2019shaping}.

\subsection*{The evolution of emotional sentiment across a sentence}

Next, we explore how the predicted distributions evolve throughout a sentence. As paradigmatic examples, we look at the case of the conjunction “but” and common intensifiers, such as “really” and “extremely”.

\subsubsection*{Conjunctions that reverse expectations}

\begin{figure}[ht]
\centering
\includegraphics[width=\linewidth]{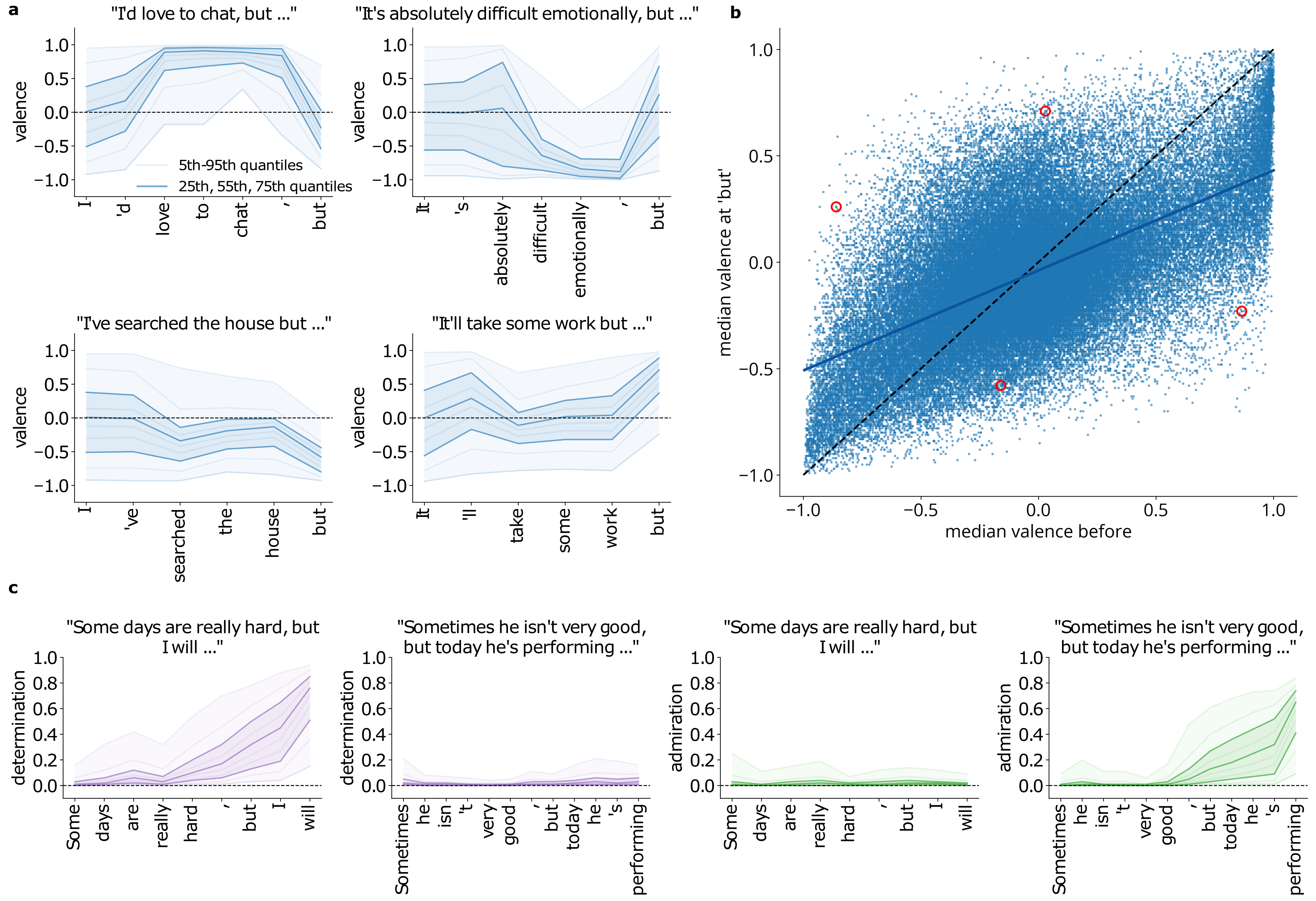}
\caption{\textbf{The expectation-reversing effects of the conjunction “but”.}  (a) Predicted quantile trajectories for four partially completed sentences show how the word “but” reverses the sentiment for preceding text, which can be of positive, negative or neutral valence. (b) The predicted valence for the preceding text (x-axis) and for the word “but” (y-axis) are shown for these four examples (red encircled points) against a backdrop of validation sentences. The significant deviation of these points away from the identity line shows the prevalence of this sentiment-reversing effect. To confirm that the transitions from neutral to positive or negative valence were also accurate (i.e., the cloud of points at center), a calibration analysis (showing a maximum absolute error of 1.8\%) was performed. (c) The word “but” can also anticipate more subtle emotions, such as determination and admiration, depending on the preceding context.}
\label{fig:but}
\end{figure}

The typical function of the word “but” is to present contrast or exception. For the case of emotions, we are also all too familiar with its role in suddenly reversing our expectations, as in the example, “I'd love to chat, but …”. In Figure~\ref{fig:but}a, we show this effect brought to life, for four different sentences. In two cases (top row), the predicted valence rises (or dips) to an extreme value before swinging to the opposite valence at the word “but”; with the outermost quantile (5th or 95th), however, anticipating this potential turnabout. For the other two cases, the valence starts at nearly the same neutral level prior to the word “but”, but then rises or falls depending on the preceding context.

To show this effect on aggregate, we plot the median predicted sentiment at the word “but” against the preceding sentiment, for the validation sentences (Figure~\ref{fig:but}b; the four red encircled points are the examples from panel a). A significant deviation away from the identity line shows how often these sentiment reversals occur. These data also reveal that many occurrences of the word “but” are characterized by a transition from neutral to either positive or negative valence (like the bottom two examples in panel a). To show that these model predictions are also accurate, we performed a separate calibration analysis for the quantiles of “but”. This had a maximum absolute error of 1.8\% across quantiles, confirming the accuracy of these predictions.

The word “but” can also signal more subtle shifts in the expression of other emotions. We demonstrate this again by way of example, in Figure~\ref{fig:but}c. Here, the word “but” signals rising determination for a sentence starting with hardship (e.g., “Some days are really hard, but …”) and rising admiration when the preceding text is socially critical (e.g., “Sometimes he isn’t very good, but …”). The reverse predictions, however, are not made, pointing to the specificity of the models, even for two related and positive emotional sentiments.

\subsubsection*{Words that widen the distribution}

\begin{figure}[ht]
\centering
\includegraphics[width=\linewidth]{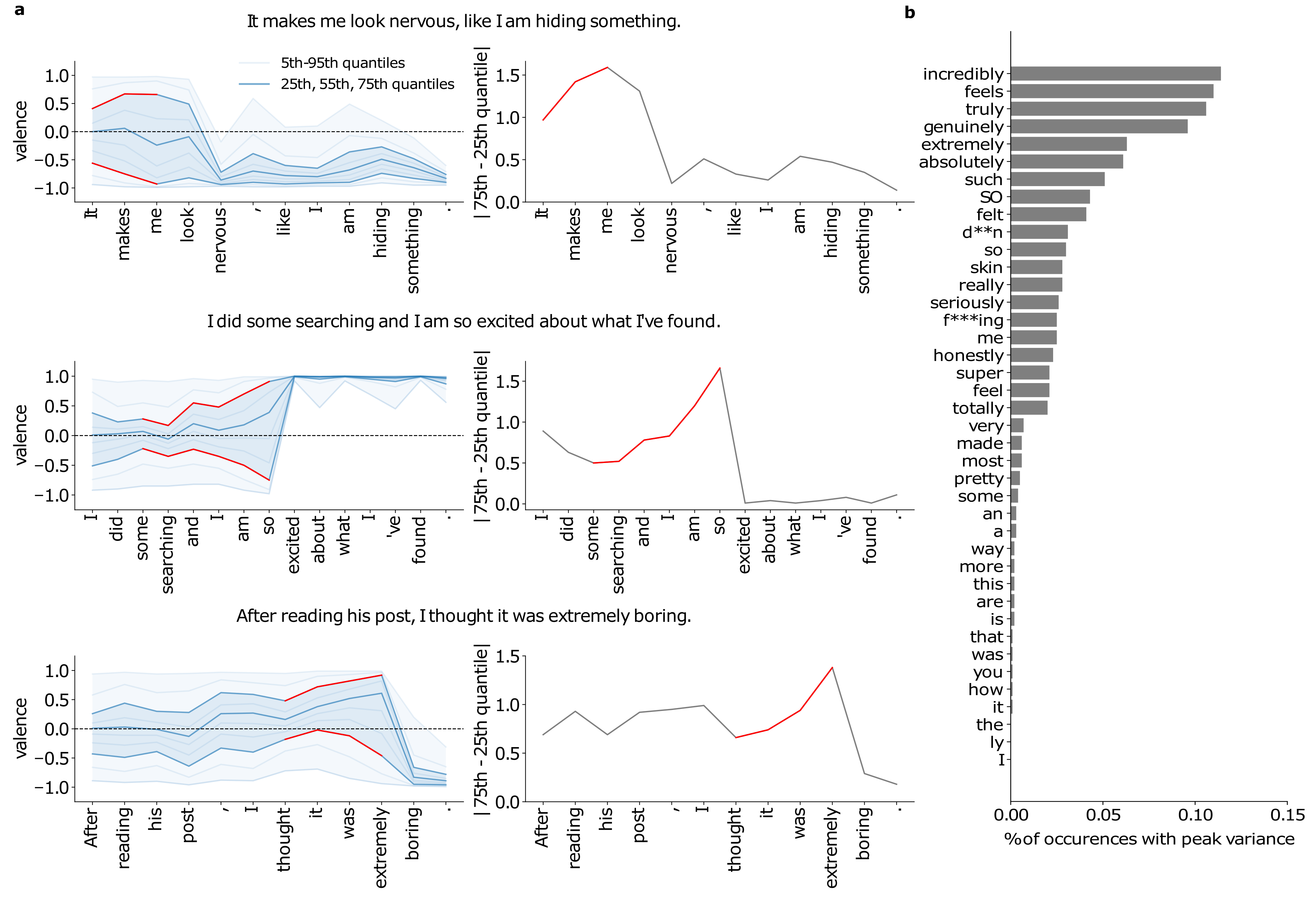}
\caption{\textbf{Variance in the distribution of sentiment across tokens.} (a) Sentences were scanned for locations at which the variance peaks (measured as the absolute value of the difference between the 25th and 75th quantiles), revealing words and phrases that predict strong changes in sentiment but do not indicate the direction of change. (b) The 40 words that occur most often at these peaks in variance include many textbook intensifiers, such as “so”, “really”, and “extremely”, as well more colloquial ones, such as “honestly” and “genuinely”; these words are sorted by the percentage of their (overall) occurrences that are associated with peak variance. These context-sensitive predictions for variance could enhance traditional lexicon-based analyses for the use of extreme language.}
\label{fig:very}
\end{figure}

A benefit of predicting the whole distribution of emotional sentiments, rather than simply their expected value, is that one can analyze differences in the level of uncertainty about how a sentence might end. Thus, we next turn to identifying locations in a sentence at which the variance peaks -- that is, places where the sentiment is poised to become more extreme, but the direction of change is not yet known. We used the absolute value of the difference between the 25th and 75th quantiles as a measure of variance, and scanned the validation sentences for local maximums. Three examples are shown in Figure~\ref{fig:very}a, with the peaks in variance and the phrases that precede them highlighted in red. For each of these examples (“It makes me …”, “I am so …”, and “it was extremely …”), it is clear that a judgment is pending, but what that may be is completely opaque. After the variance and the envelope of quantiles expands, it typically collapses, and the predicted valence drops or rises as the ambiguity around sentiment is resolved.

Context-sensitive predictions for variance could also be used enhance traditional lexicon-based analyses for the use of extreme language, which might simply count the occurrences of common intensifiers, such as the words “really” or “extremely”. Tokens associated with peak variance in our analysis (defined as the top 1\% of maximum variances across sentences), indeed include these words and many others (Figure~\ref{fig:very}b). However, this set also includes more colloquial intensifiers, such as “honestly”, “pretty”, and “seriously”, whose role as such will likely strongly depend on context. In Supplemental Table \ref{table:supp-wiki}, we list common English intensifiers from Wikipedia\cite{wiki_intensifiers} and statistics associated their predicted variances in the validation set. Roughly half of these words occur at peaks in variance (i.e., top 1\% of variances), and the other half have either only slightly lower maximal variance or occur very infrequently in our data.

\FloatBarrier

\subsection*{Generating text from the tails of a distribution}

Having demonstrated the predictive capabilities of the quantile models, we next turn to whether these can be used steer a language model to write in a way that is more positive, negative, or emotionally colored.

Generating text from a large language model operates, in its most basic form, by mapping partially completed text to a probability distribution over possible next tokens and then sampling one of these tokens. We intervene on this process by adjusting these probabilities, so that tokens that predict (and engender) certain emotions are more likely to be selected. This means that if we want, for instance, to generate text from the \emph{lower} $\alpha$-quantile (which we denote by $\alpha^-$) of a prompt's distribution, we would increase the probability for tokens that have a more negative set of predicted quantiles and decrease the probability for tokens that have a more positive set. More specifically, we re-weight the next-token probabilities proportional to the amount of each next token's predicted distribution (i.e., the number of predicted quantiles) that falls below the target $\alpha$-quantile. Sampling from these re-weighted probabilities then provides a principled way of generating sentences that reside within the lower $\alpha$-tail of the prompt (for more details, see Methods). To generate sentences \emph{above} the $(1-\alpha)$-quantile (which we denote by $\alpha^+$), we reverse the values of the quantiles.

\begin{figure}[ht]
\centering
\includegraphics[width=\linewidth]{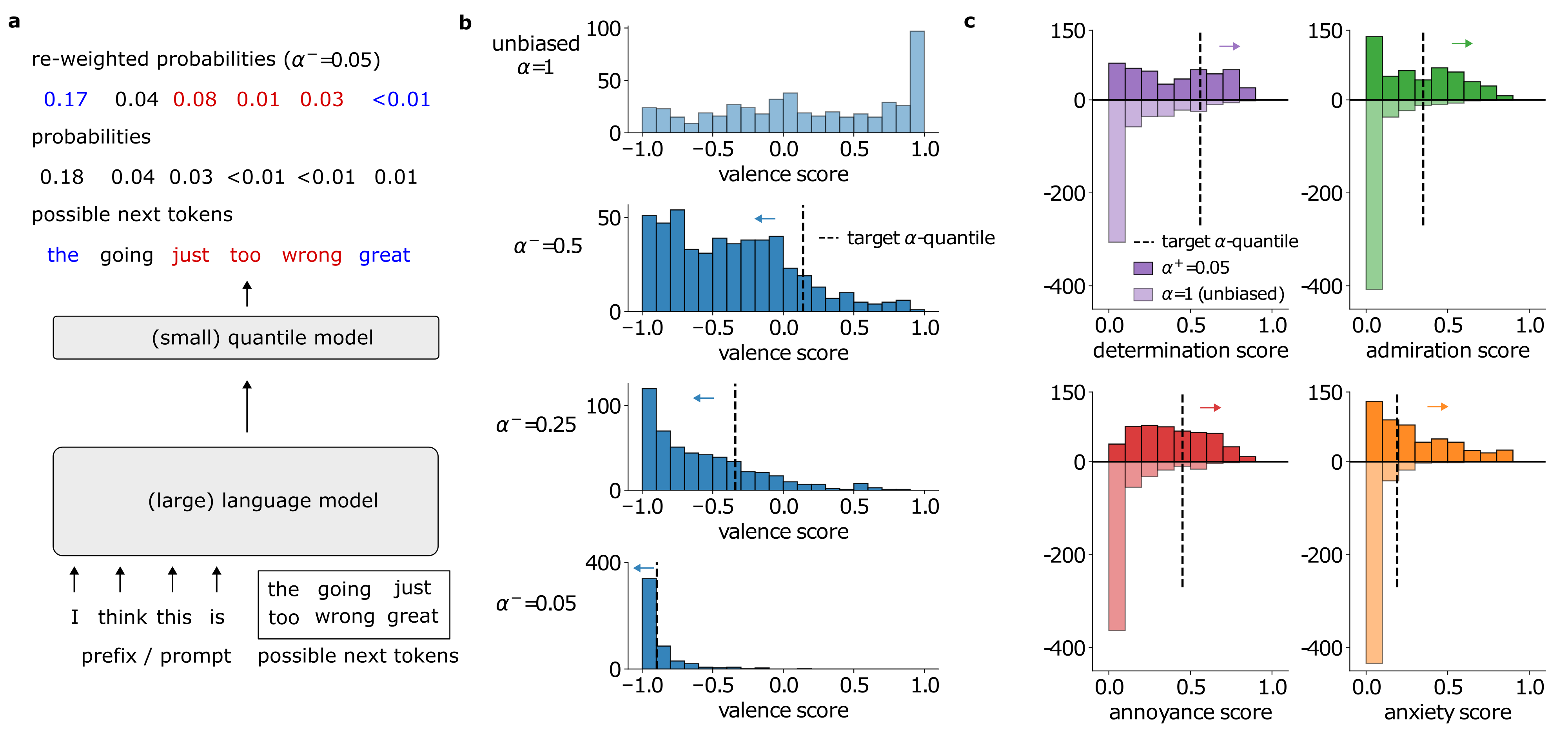}
\caption{\textbf{Steering a large language model to write more emotionally.} (a) We intervene on the normal text generation process (of a large language model) by re-weighting the next token probabilities using their predicted sentiment quantiles. Tokens more likely to produce sentences below a target quantile are upweighted (shown in \textcolor{red}{red}) and tokens less likely to produce sentences below that quantile are downweighted (shown in \textcolor{blue}{blue}). (b) Setting the target quantile to lower values of $\alpha^-$ (which corresponds to the lower tail) produces a higher concentration of negatively valenced utterances than the unbiased language model ($\alpha=1.0$); more positively valenced text can also be produced by reversing the applicable tail for $\alpha$ (not shown here). The black dashed vertical lines correspond to the target $\alpha$-quantiles calculated from unbiased language model's distribution; note that most of the generated sentences fall below the target for each $\alpha^-<1$.  (c) Using the four emotion models and a target of $\alpha^+=0.05$ for the \emph{upper} tail produces text with much higher scores along each respective emotional dimension relative to the unbiased model (shown inverted and underneath, $\alpha=1.0$). Each distribution contains 500 generations, aggregated across five different prompts.}
\label{fig:genscheme}
\end{figure}

We provide a schematic example of this procedure in Figure~\ref{fig:genscheme}. Here, the initial $\alpha^-$, for the prompt “I think this is”, is set to 0.05 (to sample from the extreme lower tail). The next possible tokens include “the”, “going”, “just”, “wrong”, “too”, and “great”, and each is assigned a probability by the language model. To re-weight these probabilities, we look at the estimated quantiles for each next token, and upweight/downweight those tokens depending on if they have more/fewer of their quantiles (i.e., more/less probability mass) below the $\alpha^-$-quantile of the prompt's distribution. “Wrong”, “just” and “too” are upweighted because they have more negative valence distributions, while “great” and “there” are downweighted because they have more positive distributions. The next token is then sampled according to the re-weighted probabilities and the process is repeated.

To demonstrate the effectiveness of this method, we use the trained quantile models along with GPT-2 to complete five prompts (“I think this is”, “This week is really busy,”, “I”,  “Yesterday I went to school and”, “My kids”) with varying degrees of negativity or emotional tones. In Figure~\ref{fig:genscheme}b, we show how lowering $\alpha^-$ generates sentences from increasingly negative lower tails of the valence distribution. For values of $\alpha^-$ less than 1, the majority of the generated distribution lies below the target quantile (vertical black dashed line) with only a small percentage of sentences with scores above that value. In Figure~\ref{fig:genscheme}c, we set $\alpha^+=0.05$ (for the upper tail) and sample according to each of the emotional models; here, we can see that the distributions are strongly shifted to the right relative to the distribution of sentences generated from an unbiased process (shown inverted and underneath; $\alpha=1.0$); however, sampling sentences from an extreme quantile for these sparse emotion distributions proved to be more difficult than for valence, with the quantile targeting being imperfect.

\begin{figure}[ht]
\centering
\includegraphics[width=\linewidth]{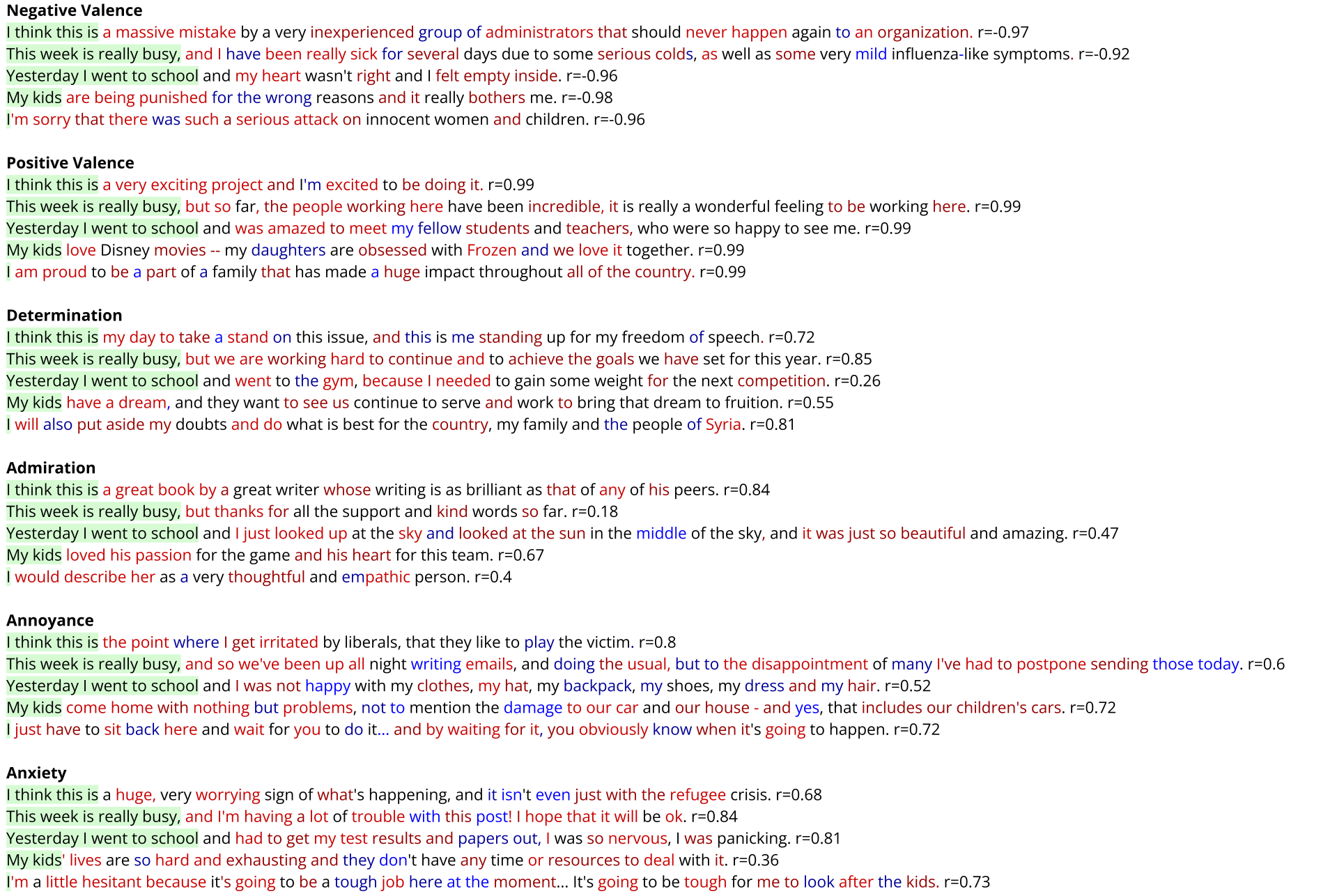}
\caption{\textbf{Examples of emotional text generated.} Color is used to show re-weighted token probabilities; bright red indicates upweighting by more than 1.5x (dark red means greater than 1.1x) and bright blue means downweighting by less than 66\% (dark blue means less than 80\%). The prompt is highlighted in green. Sentences for negative valence were sampled from the lower tail ($\alpha^-=0.05$), while sentences from other five emotional categories are sampled from the upper tail ($\alpha^+=0.05$). These examples were handpicked to show the linguistic diversity in what the model chooses to reweight. However, more examples for each sentiment are shown in Supplemental Figures~\ref{fig:supp-gen-val-1}-\ref{fig:supp-gen-det-adm-05}.}
\label{fig:genexamples}
\end{figure}

In Figure~\ref{fig:genexamples}, we show example sentences from each of the models (with $\alpha^{-}$ or $\alpha^{+}$ set to 0.05) and color the text to show the re-weighted probabilities for each token. As expected, the model chooses to upweight words that have a clear valence or emotional tone, such as “sorry” and “mistake”, and at the opposite end of the spectrum “exciting” and “amazed”. However, because the model is predictive, it also upweights words and phrases that are less obviously valenced, but which make the target sentiment or emotion more likely to occur. Some of these are predictive because of their semantics, such as “test results” for anxiety, “sky” for admiration or “gym” for determination, and others for their grammatical role, such as “not”, “and”, and “very”. Indeed, the word “but” plays a critical role in the generation of the positive emotions. Finally, there are also phrases, such as “this is my day” for determination or “this is the point” for annoyance, which convey emotions idiomatically (and subtly), but are nevertheless able to be exploited by the model. More examples are shown in Supplemental Figures~\ref{fig:supp-gen-val-1}-\ref{fig:supp-gen-det-adm-05}.

\FloatBarrier

\section*{Discussion}

In this paper, we demonstrated how large language models can provide a novel lens onto the inner structure and emotional dynamics of sentences. We developed a method for training a model to predict, word-by-word, the quantiles of a distribution of the final emotional sentiments of text, and validated these predictions on unseen sentences. We then showed how the trajectories of the predicted quantiles capture the intuitive emotional effects of conjunctions (i.e., “but”) and intensifiers (e.g., “really” and “so”). Finally, we used the quantile models to encourage GPT-2 to write more emotionally colorful sentences from specific quantiles of a distribution. 

Training quantile models and using them to generate text is related to a number of recent trends in machine learning and natural language processing. Predicting a distribution rather than a single value for sentiment is related to distributional reinforcement learning, which estimates distributions of future rewards in addition to the expected value and then optimizes for some aspect of that distribution; indeed, using quantile regression to predict end scores can be viewed as a Monte Carlo (also known as TD(1) in the reinforcement learning literature\cite{sutton2018reinforcement}) version of the learning algorithm used in previous work\cite{dabney2018distributional, bellemare2023distributional} and sampling tokens from the re-weighted probabilities can be seen as a basic form of risk-“aware” control. For quantile estimation, we found that Monte-Carlo methods produced better calibrated models than temporal difference methods such as TD(0). We speculate that this may be due to the need of TD(0) to pass predictive information backwards through every token, which could be susceptible to tokens with lower quality hidden state representations. For control, we showed how a language model can be steered to generate from either the lower or upper tail of a distribution by reweighting next-token probabilities according their predicted quantiles. However, it is also possible to optimize for other aspects of the predicted distributions, for instance, generating in a way that avoids (rather than targets) the lower tail. This could be useful for reducing the amount of toxic content, when paired with a toxic comment classifier\cite{hanu2020unitary}. A simple approach for this would be to re-weight next token probabilities according to 1-$\alpha$ instead of $\alpha$, though future work would needed to explore this and other risk-sensitive approaches to text generation more fully.

Our method for text generation is also related to several recent methods from NLP that re-weight the next-token probabilities to encourage language models to generate text with particular characteristics. For example, \cite{krause2020gedi, holtzman2018learning, yang2021fudge} use lightweight discriminative classifiers to heuristically bias the next-token logits. Other work \cite{snell2022context, snell2022offline, guo2021text, leblond2021machine} estimates Q-value functions, which like our quantile models are forward-looking, to steer text generation (again via probability re-weighting). However, none of this prior work has tried to estimate distributions of future characteristics (i.e., sentiment) and target particular quantiles (which is why we do not compare directly  with these methods); moreover, many of the probability re-weighting techniques are heuristic, as opposed to the principled re-weighting scheme we develop. This also contrasts with prompting, which while often effective at generating text with certain characteristics, is largely conducted on a trial-and-error basis.

Other methods for controlled text generation involve re-training large language models. One well-known approach is to use control codes during pretraining\cite{keskar2019ctrl} (or finetuning \cite{korbak2023pretraining}) to write in different styles or about different topics. There is also a recent trend of reinforcement learning with human feedback (RLHF)\cite{stiennon2020learning, ouyang2022training}, which uses a model trained on human feedback to provide `end' scores, and using finetuning via policy gradient to optimize for these scores. Predicting end scores via a separate (quantile) model and probability re-weighting can be seen as a model-based variant of this approach. Finally, the most closely related work is \cite{lu2022quark}, conducted contemporaneously, which uses finetuning and conditioning on a special tokens to generate from one of five quantiles. An interesting future path would be to explore the differences between conditional pretraining or finetuning and online probability-reweighting methods, such as ours. The latter involve more computation at inference but avoid the need to re-train large language models, which is becoming increasingly difficult to do as the scale of these models increases. Using smaller models to guide larger ones also allows for easier mixing and matching between control tasks and LLMs, for instance using BERT (110M parameters) to predict the accuracy of BlenderBot (2.7B parameters) and reduce the number of over-confidently expressed falsehoods\cite{mielke2022reducing}. More work is needed, however, to know the domains in which smaller models can be used to guide larger ones effectively (and the domains in which richer representations of larger models are truly needed).

Generation aside, predictive quantile models  may be useful for analyzing text in a wide variety of ways, complementing the long tradition of lexical analysis, use of word-embeddings, and more recently the use of contextualized embeddings. Our analyses of conjunctions (i.e., “but”) and intensifiers (e.g., “really”, “so”) are only a starting point. One could search for collections of phrases with unique emotional dynamics using unsupervised methods, such as time-series motif analyses or clustering. Features of the emotional quantile trajectories, such as variance, could also be used to augment more traditional lexicon-based analyses\cite{chung2007psychological}, in, for instance, quantifying extreme language. They could also provide insights into the workings of humor. Word-by-word level features, like variance, and aggregate features, like the $\alpha$-quantiles that individuals favor, can also be used to compare writing from different sources. In the Supplemental Information, we provide an example the latter by inverting our text-generation process to infer the most likely alpha value for each sentence. Averaging alpha values across sentences, we observe a slight pessimism bias in posts made by Reddit users who self-report depression relative to those from healthy control users. We also use this method to score excerpts from well-known speeches and books, highlighting the dominant emotion in each text (e.g., determination in the "We shall fight on the beaches" speech delivered by Winston Churchill, Figure \ref{fig:alphaanalysis}). Analyses like these, as well as those based on other aggregate features, including the posterior distributions over the values of
alpha, could be used for the sake of understanding the style of individuals (famous writers, or newspaper outlets), analyzing historical trends \cite{card2022computational}, or identifying the characteristics of effective therapeutic language \cite{miner2022computational}. Moreover, predictive models could be estimated for other semantic characteristics, such as a forward-looking version of the topic classifier used to analyze the narrative arc of short dialogs or scripts\cite{mathewson2019shaping}. Note that forward-looking models (like the ones we explore here) differ from classification methods (which view the whole text), and could be an alternative to feature importance methods such as attention-weight analysis, SHAP\cite{lundberg2017unified, shapley1953value} or LIME\cite{ribeiro2016should} for highlighting important words; moreover, forward looking models reveal how predictions for end-scores change as information is added, mirroring the dynamics of human expectations as we read.

A relevant (albeit difficult) next direction would be to develop methods for predicting the emotional arc of text at longer time-horizons -- conducting the same sort of analyses and generations at the scale of paragraphs, sections, chapters and beyond. Relatedly, one could attempt to predict the longer-run destination of a person's thought (or at least the sentiment thereof) based on the words so far said. Although this would be undoubtedly difficult for most situations, there may be contexts in which repeatable patterns might be expected, e.g. in the context of psychotherapy. If successful, this could be used to flag cognitive biases \cite{beck1979cognitive, al2018absolute} (e.g., use of absolute language  like “always”, etc.), which although they are relatively neutral in themselves, can nevertheless lead perniciously to more extreme (and often) negative thoughts. This could be done real-time during telehealth therapy sessions or retrospectively as patients and therapists review previous conversations.

In recent work\cite{gagne2022peril}, we also explored a connection between worry and risk aversion in an abstract (and very simplified) computational model, which presented worry as a form of model-based (offline) planning aimed at mitigating (bad) outcomes from the lower $\alpha$-quantile of a distribution. In this model, the process of worrying involves upweighting the probabilities (in an internal cognitive model) for the lower $\alpha$-tail of the possible outcomes and the states that precede them, and then designing contingency plans to avoid these scenarios. Since worry is often thought to be mediated by verbal thought (i.e., language), it would be useful to develop a verbal version of this model, using a large language model and next-token probability weighting, to model phenomena such as the forms of catastrophizing evident in 'what if' chains-of-thought\cite{davey1998catastrophic}.

One major limitation of our work is that we use a state-of-yesterday's-art language model (GPT-2 large), which pales in comparison to the recent and much larger, but unfortunately inaccessable, models such as Chinchilla\cite{hoffmann2022training}, Lambda\cite{thoppilan2022lamda}, and ChatGPT\cite{ouyang2022training}. Indeed, generation from GPT-2 (even without re-weighting) tended to produce run-on sentences and flaunt standard grammatical convention (e.g., using comma splices). The quantile models were also sometimes `jumpy' for rare or unusual tokens or text, despite the very good marginal calibration for common prefixes. A second limitation is that the quantile models rely on the underlying scoring algorithms applied to the dataset employed. For some of the emotions that we did not report here, the algorithm was rather shallow, overly emphasizing a small subset of words. Furthermore, the Reddit dataset appears to be a poor source for capturing the range of certain emotions. Finally, both the end sentiment model and emotion classifier were trained to predict the probability of the text containing that sentiment or emotion (as judged by a human rater) and therefore do not directly capture the intensity of emotional expression. Nevertheless, our method and demonstrations stand as a proof-of-principle, and we expect only improvements with models of increased scale and predictive capabilities.

\section*{Methods}

\subsection*{Datasets and text preprocessing}

Text used for training the quantile models was combined from several sources: SMHD \cite{cohan2018smhd}, Social IQ \cite{zadeh2019social}, Positive Psychology Frames \cite{ziems2022inducing}, and sentences generated from GPT-2 \cite{radford2019language}. The largest source was the SMHD dataset, which contains posts to Reddit between January 2006 and December 2017 (inclusive). A random subset of posts from the full dataset was sampled, split into sentences, and then grouped into a training set (2M sentences) and a validation set (1M sentences). The SMHD also includes labels for each Reddit user for whether they self-reported a psychiatric diagnosis (e.g., depression) or whether they were recruited as a healhy control. We also included 30k sentences from the Social IQ dataset, 13k sentences from the Positive Reframes dataset, and 60k sentences generated from GPT-2 in response to short prompts, such as “I woke up early.”, “I made dinner” etc. These additional sentences were added out of convenience and originally intended for analyses unrelated to the current work. They were also divided into train and validation sets.

Sentences were extracted from longer segments of text using Spacy’s NLP “sentencizer”\cite{spacy2}. Sentences that were fewer than 5 words or those contained special punctuation characters, such as pipes (“|”), new line characters (“\textbackslash r”, “\textbackslash n”), colons, parentheses, and brackets were excluded. Sentences were therefore only punctuated by a period, a question mark or an exclamation mark. This was done to limit the number of run-on utterances in the dataset.

Prior to training the quantile models, the data was tokenized using the GPT-2 tokenizer implemented by Huggingface and sentences were truncated to 32 tokens for computational reasons; this resulted in only about 7\% of the sentences being truncated. Again for computational efficiency, the token embeddings for each sentence were extracted from the final layer of GPT-2 large and saved to disk prior to fitting the quantile models.

\subsection*{Pretrained models}

The large version of GPT-2\cite{radford2019language} (with 774M parameters), implemented using code from Huggingface, was used to extract text embeddings and to generate text. Positive and negative valence scores were provided by a version of RoBERTa finetuned on tweets from the TweetEval sentiment dataset \cite{barbieri2020tweeteval}. This model outputs separate probabilities for negative and positive valence, however we combined these by multiplying by -1 and 1 (respectively) and summing to form a single scale, which varied continuously between -1 to 1. Scores for the other emotions were provided by a version of BERT that was finetuned on short English comments (3-30 tokens long) posted to Reddit and rated by human annotators for emotional content. The emotion model, provided by Hume AI, is an improved version of that used in previous work \cite{demszky2020goemotions}, retrained using approximately 5x more data and roughly double the number of emotional categories (53 instead of 27). Scores ranged between 0 to 1 for each emotion and represent the probability that a human rater would label a piece of text with that emotion. The top two principal components of these emotional scores correspond approximately to the dimensions of ‘valence’ and ‘arousal’ that are central to the circumplex model of affect \cite{russell1980circumplex, russell1999core, russell2003core, barrett2004feelings} (Supplemental Figure \ref{fig:pca}). For our analyses, we focused on two positive emotions (determination and admiration) and two negative emotions (anxiety and annoyance). These were selected based on the linguistic diversity of the sentences receiving high scores for that emotion.

\subsection*{Quantile models}

Separate quantile models were used to predict the distributions of \emph{end} scores for valence, determination, admiration, annoyance, and anxiety. For each token $x_t$ in a sentence, a quantile model  outputs ten equally-spaced quantiles between $\alpha=0.05$ and $\alpha=0.95$. The model's predictions, denoted by $q_\alpha(x_t|x_{<t}; \theta)$, are also conditioned on the preceding text $x_{<t}$ through the hidden states of the large language model, which are provided as input.

Each quantile model consisted of single hidden layer with 100 hidden units (whose parameters are denoted by $\theta$). The input dimension was 1280, corresponding to GPT-2 large’s hidden state size for a single token, and the output dimension was 10, corresponding to the number of quantiles used (5\%, 15\%, \ldots, 95\%). Each network had 130k parameters, which is substantially smaller than GPT-2 (at 774M parameters). A tanh or sigmoid transformation was applied to the output depending on if the model was being trained on positive and negative valence ([-1,1]) or emotions ([0,1]).

Given a sentence $x$ and an end score $y$ sampled from the training dataset, the parameters $\theta$ for a quantile model were updated by performing gradient descent on the following \emph{Huber quantile loss} \cite{Koenker2005, dabney2018distributional, bellemare2023distributional}:

$$ L(\theta) = \sum_\alpha \sum_t \rho^H_\alpha \bigl( y - q_\alpha(x_t|x_{<t}; \theta) \bigl), \;\;\;\;\;\;\;     \rho^H_\alpha(u) = |\mathbbm{1}_{u<0} - \alpha | H(u) $$

Where $\rho^H_\alpha$ is an asymmetric weighting function that penalizes underestimation errors ($y>q_\alpha$) with a weight of $\alpha$ and overestimation error ($y < q_\alpha$) with a weight of $1-\alpha$, and $\mathbbm{1}$ is an indicator function which takes on the value  1 if the condition in the subscript is true and the value  0 otherwise. To make the loss smooth at zero, the function $H$ is applied to the estimation errors, where $H(x)$ is $0.5x^2$ if $|x| \le k$ and $(|x|-0.5k)k$ otherwise. We set $k$=0.001.

Batches of 20 sentences were used to train the model, with the order of the sentences randomized at the start of each epoch. The models were trained for 25 epochs and evaluated periodically on a 100k-sentence subset of the validation dataset. Adam\cite{kingma2014adam} was used with an initial learning rate of 1e-4; the learning rate was also decayed by half each time the validation loss plateaued for two consecutive epochs. The quantile models at the end of training were used for all analyses. The training and validation curves are plotted in Supplemental Figure~\ref{fig:training_loss}.

\subsection*{Temporal difference methods}

As an alternative, we also experimented with using temporal difference methods, specifically TD(0), to train the quantile models \cite{dabney2018distributional, bellemare2023distributional}. Instead of using the end scores as the target for the quantile predictions of each token, the predicted distributions for the next token were used. This was accomplished by substituting $q_{\alpha^{\prime}}(x_{t+1}|x_{<t+1}; \theta)$ with $y$ in the equation for the quantile loss and additionally summing over all pairwise combinations of $\alpha^\prime$ and $\alpha$ (see \cite{dabney2018distributional, bellemare2023distributional} for more details). Note that using quantile regression directly on the end scores $y$ can be viewed as a Monte Carlo (TD(1)) version of distributional temporal difference methods \cite{bellemare2023distributional}. However, using this method resulted in quantile models that were more poorly calibrated, with an maximum absolute error of ~9\% for valence.

\subsection*{Model calibration}

1M validation sentences were used for global calibration. For each sentence, the end score (valence or emotion) was compared to the predicted quantiles at each token position. If the end score fell below a predicted quantile, a 1 was stored, otherwise a 0 was stored. Averaging across all sentences gives the empirical probability that the end score falls below the quantile of a certain level. An accurately estimated $\alpha$-quantile would mean that the end scores fell below that quantile $\alpha$-\% of the time. Quantiles were calibrated separately for each token position; the number of sentences used for each position differed, because sentences shorter than the token position analyzed were naturally excluded. The results of the calibration are shown in Supplemental Figure~\ref{fig:calibration}.

\subsection*{Generating text for specific $\alpha$-quantiles}

Text generation in many large language models, like GPT-2, is autoregressive. Given a partially completed sentence $x_{< t}$ (and prompt $x_p$), the LLM outputs a probability distribution over possible next tokens $p(x_t|x_{< t}, x_p)$. A single token $x_t$ is sampled from this distribution and is then fed back into the model as input for the next step. 

To generate more positively or negatively valenced text, or text with a specific emotional tone, we reweight these next-token probabilities using the (valence or emotion) quantiles predicted for each token. More specifically, our goal is to generate sentences from below a prespecified $\alpha$-quantile of a prompt $x_p$'s distribution. To do so, we adjust the next-token probabilities to be:

$$p^\prime(x_t|x_{< t}, x_p) \propto \alpha_{x_t}p(x_t|x_{< t}, x_p)$$ 

Where the weight $\alpha_{x_t}$ corresponds to the proportion of each next token's predicted distribution that resides below the $\alpha$-quantile of prompt's distribution $q_\alpha(x_p)$. These weights are estimated by linearly interpolating the value for $\alpha_{x_t}$ such that its corresponding  quantile equals the target quantile, i.e. such that  $q_{\alpha_{x_t}}(x_t | x_{< t}, x_p) = q_\alpha(x_p)$ for each next token. Normalization is also applied to ensure that the reweighted probabilities sum to one. For intuition, consider a next token whose predicted distribution is entirely above the target quantile. This would be assigned a weight of 0, because it does not contribute to the lower $\alpha$-tail of the prompt's distribution, and selecting it would make generating sentences from that lower tail impossible (given perfectly accurate predictions). Conversely, a token whose predicted distribution is entirely below the target quantile (i.e., within the target area) would be assigned a weight of 1 (and its probability would be unaltered). To generate sentences from above the $(1-\alpha)$-quantile, the same procedure is applied to quantiles that are multiplied by -1. We denote the lower tail using $\alpha^-$ and the upper tail using $\alpha^+$.

After the next token is sampled from the reweighted probabilities, the process is repeated by comparing the distributions for the next set of possible tokens to the target quantile of the prompt distribution. Probability re-weighting is done after truncating the distribution using top-p (with p=0.95) and top-k (k=50) sampling. This was done to reduce the probability that extremely rare or odd tokens are selected and to increase computational speed. The temperature parameter was set to 1.

\subsubsection*{Further details for text generation}
\label{sec:gen-details}

Text was generated from the model until a period token was chosen or a maximum sequence length of 40 tokens was reached. For both the unbiased and biased models many generation runs failed to self-terminate before the maximum sequence length. For the unbiased GPT-2 model ($\alpha=1.0$), 79/500 utterances did not terminate. For the negative-valence model, this number was 77/500 (for $\alpha^-=0.5$), 113/500 (for $\alpha^-=0.25$), and 111/500 (for $\alpha^-=0.05$). For the positive-valence model it was 79/500 (for $\alpha^+=0.5$), 82/500 (for $\alpha^+=0.25$), and 74/500 (for $\alpha^+=0.05$). For the emotion models at $\alpha^+=0.05$, it was 67/500 (determination), 79/500 (admiration), 91/500 (anxiety), and 130/500 (annoyance). 

In general, GPT-2 size language models are known to have issues with repetition and run-on sentences \cite{holtzman2019curious, xu2022learning}. This issue is likely also exacerbated by using very simple prompts, which do not indicate how long the sentence should be. However, given that the quantile models were trained on short sentences (<32 tokens), we opted for shorter prompts to keep the quantile models in-distribution. Aside from using a larger language model, this problem could also potentially be mitigated by longer prompts or adjusting the temperature parameter. As the current work aims merely to demonstrate the validity of this approach, we will leave these explorations for future work.

\section*{Data availability}

The primary dataset we used to train our models, SMHD \cite{cohan2018smhd}, was obtained from Georgetown University with permission to use but not to share, in order to protect user privacy. We therefore cannot make this dataset available. However, scripts used to process the data and train the models will be included in the project’s Github repository.

\section*{Acknowledgements}

CG and PD are funded by the Max Planck Society. PD is also funded by the Alexander von Humboldt Foundation. The authors would like to thank Alan Cowen and Hume AI for kindly providing access to the language model that was used to emotionally score text. We would also like to thank Christian Adam, Y-Lan Boureau, Marc Bellemare, and Marcel Binz for helpful comments on an earlier version of this manuscript.

\section*{Author contributions statement}

C.G. and P.D. conceived the experiments and analyses; C.G. conducted the experiments; C.G. and P.D. wrote and reviewed the manuscript. 

\section*{Additional information}


\textbf{Competing interests:} The author(s) declare no competing interests.

\bibliography{main}

\clearpage
\section*{Supplemental Information}

\subsection*{Inferring the $\alpha$ for a given text}

For generating text, we specified a particular $\alpha$-quantile below (or above) which to sample sentences. In this section, we explore a method for inverting this process and inferring the the most likely $\alpha$ that would have produced a given sentence or a set of sentences written by a single source.


\subsubsection*{Estimating $\alpha$ for single sentences}
To estimate $\alpha$, we follow the same procedure as generation, except that we yoke the next tokens to those in a sentence. As the model processes that sentence, we separately sum the tokens’ log-probabilities under the language model and the reweighted log-probabilities under different values of $\alpha^-$ and $\alpha^+$. The alpha value with the highest summed log-probability is considered the best fitting for that sentence (including $\alpha=1.0$, which corresponds to the baseline language model).

Under different alphas, token probabilities are reweighted by different amounts, with valenced words being more strongly up- (or down-) weighted by the lowest values of alpha. Therefore, highly valenced sentences will be most likely under the lowest values of alpha, and ambivalent sentences will be best fit by a value of alpha closer to 1. In Figure~\ref{fig:alphaanalysis}a, we show two example sentences, both with negative valence. For the first sentence, negative words are primarily upweighted, and the rest untouched, leading to a best fitting $\alpha^-$=0.1. For the second, which is more ambivalent, the positive part of the sentence is too unlikely under the extreme $\alpha^-$'s (<=0.1), leading to a best fitting $\alpha^-$=0.25.

The final valence of the two sentences is -0.97 and -0.66, respectively, in an order that conforms with the best fitting values of alpha. However, the posterior distributions over $\alpha^-$ (or, here, the likelihood curves, assuming a flat prior) provides a finer-scale depiction of the emotional workings of the sentence, showing, for instance, how very unlikely it is that the second sentence was associated with a really low value of $\alpha^-$ (presumably because of the change in sentiment of the end clause), whereas the first sentence has a far flatter curve.

\subsubsection*{Generation and recovery analysis}

We also performed a generation-recovery analysis to assess the accuracy with which we can infer the alpha value for a single source (e.g., an individual or another LLM). To have a ground truth against which to validate, we generated 500 hundred sentences for each $\alpha^-$ and $\alpha^+$ value from a larger set of possible values (i.e., [0.05, 0.1, 0.25, 0.5, 0.75]) and treated each set of sentences as a single source. Averaging the best-fitting $\alpha^-$ or $\alpha^+$ across sentences produced an almost entirely correct re-estimation of the alpha value (i.e., the one used to generate that set of sentences; Pearson correlation > 0.99; Figure~\ref{fig:alphaanalysis}b). Note that before averaging, the $\alpha^+$ values were converted to [1.25, 1.5, 1.75, 1.9, 1.95], so that they could be differentiated from the $\alpha^-$ values.

We then repeated this analysis for different sample sizes from 1 sentence to 500 sentences per source (with 50 simulations per sample size) and show that even with 50 sentences, the ground truth $\alpha$ can be recovered 77\% of the time (compared to 90\% for 500 sentences; top left inset in  Figure~\ref{fig:alphaanalysis}b). This demonstrates that around fifty sentences suffices to estimate the statistical tendency to generate sentences from below/above a particular quantile of the distribution.

\subsubsection*{Analyzing text written by Reddit users}

As a test of external validity, we identified Reddit users who had written at least fifty sentences from our validation set and split them into two groups: those with self-reported diagnoses of depression and those who were designated as health controls. These labels were derived from the SMHD dataset\cite{cohan2018smhd}, the largest component of our training and validation sets. To balance the two groups, 192 participants were randomly selected from larger group (depression) and fifty sentences were randomly selected from each user. Estimating the best fitting $\alpha$ per user revealed a significant difference between the two groups (mean $\alpha^-$=0.92 for depression; mean $\alpha^-$=0.97 for control; t= -2.57, p=0.01, df=382; where $\alpha^-<1.0$ corresponds to pessimistic sampling and $\alpha=1.0$ corresponds to unbiased sampling). Moreover, 69\% of the users who self-reported a diagnosis of depression had a best fitting $\alpha^-$ below 1 compared to 57\% for control users. This is notable because the SMHD\cite{cohan2018smhd} excluded posts with explicit mentions of mental health diagnoses or those posted to mental health subreddits, and instead contains posts on a variety of topics, such as sports and politics.

\subsubsection*{Analyzing real world examples}

Finally, we show the same procedure applied to each emotion model and applied to real-world examples, to highlight the applicability of the method to analyzing different sorts of writing. We selected excerpts from well-known authors that we thought strongly conveyed each one of the four emotions (see below for the full excerpts). $\alpha^+$ was estimated first per sentence and then averaged across all sentences for each of the four texts. The estimated $\alpha^+$'s for each example was low specifically for one emotion, matching our intuitions (Figure~\ref{fig:alphaanalysis}c). For instance, Oprah Winfrey's eulogy commemorating Rosa Parks has an mean $\alpha^+$ of ~0.4 for admiration, and the excerpt from Winston Churchill's famous speech, “We shall fight on the beaches”, has a mean $\alpha^+$ of ~0.31 for determination. These values of $\alpha^+$ indicate the upper-tail quantile, with lower values signifying more extreme emotional scores.

For determination, we used the following speech from Winston Churchill delivered on June 4, 1940: 

\begin{quote}
I have, myself, full confidence that if all do their duty, if nothing is neglected, and if the best arrangements are made, as they are being made, we shall prove ourselves once again able to defend our Island home, to ride out the storm of war, and to outlive the menace of tyranny, if necessary for years, if necessary alone. At any rate, that is what we are going to try to do. That is the resolve of His Majesty’s Government-every man of them. That is the will of Parliament and the nation. The British Empire and the French Republic, linked together in their cause and in their need, will defend to the death their native soil, aiding each other like good comrades to the utmost of their strength. Even though large tracts of Europe and many old and famous States have fallen or may fall into the grip of the Gestapo and all the odious apparatus of Nazi rule, we shall not flag or fail. We shall go on to the end, we shall fight in France, we shall fight on the seas and oceans, we shall fight with growing confidence and growing strength in the air, we shall defend our Island, whatever the cost may be, we shall fight on the beaches, we shall fight on the landing grounds, we shall fight in the fields and in the streets, we shall fight in the hills. We shall never surrender, and even if, which I do not for a moment believe, this Island or a large part of it were subjugated and starving, then our Empire beyond the seas, armed and guarded by the British Fleet, would carry on the struggle, until, in God’s good time, the New World, with all its power and might, steps forth to the rescue and the liberation of the old.
\end{quote}

\noindent For admiration, we used the following section of the eulogy for Rosa Parks, delivered by Oprah Winfrey in October 2005:

\begin{quote}
To Reverend Braxton, family, friends, admirers, and this amazing choir:

I -- I feel it an honor to be here to come and say a final goodbye.

I grew up in the South, and Rosa Parks was a hero to me long before I recognized and understood the power and impact that her life embodied. I remember my father telling me about this colored woman who had refused to give up her seat. And in my child's mind, I thought, “She must be really big.” I thought she must be at least a hundred feet tall. I imagined her being stalwart and strong and carrying a shield to hold back the white folks.

And then I grew up and had the esteemed honor of meeting her. And wasn't that a surprise. Here was this petite, almost delicate lady who was the personification of grace and goodness. And I thanked her then. I said, “Thank you,” for myself and for every colored girl, every colored boy, who didn't have heroes who were celebrated.
\end{quote}

\noindent For anxiety, we used the following passage from the book “Kiterunner” by Khaled Hosseini: 

\begin{quote}
Panic. You open your mouth. Open it so wide your jaws creak. You order your lungs to draw air, NOW, you need air, need it NOW. But your airways ignore you. They collapse, tighten, squeeze, and suddenly you’re breathing through a drinking straw. Your mouth closes and your lips purse and all you can manage is a croak. Your hands wriggle and shake. Somewhere a dam has cracked open and a flood of cold sweat spills, drenches your body. You want to scream. You would if you could. But you have to breathe to scream. Panic.
\end{quote}

\noindent And for annoyance, we used the following passage from the commencement speech delivered by David Foster Wallace at Kenyon College in 2005 titled “This Is Water: Some Thoughts, Delivered on a Significant Occasion, about Living a Compassionate Life”:

\begin{quote}
 … And many more dreary, annoying, seemingly meaningless routines besides. But that is not the point. The point is that petty, frustrating crap like this is exactly where the work of choosing is gonna come in. Because the traffic jams and crowded aisles and long checkout lines give me time to think, and if I don’t make a conscious decision about how to think and what to pay attention to, I’m gonna be pissed and miserable every time I have to shop. Because my natural default setting is the certainty that situations like this are really all about me. About MY hungriness and MY fatigue and MY desire to just get home, and it’s going to seem for all the world like everybody else is just in my way. And who are all these people in my way? And look at how repulsive most of them are, and how stupid and cow-like and dead-eyed and nonhuman they seem in the checkout line, or at how annoying and rude it is that people are talking loudly on cell phones in the middle of the line. And look at how deeply and personally unfair this is.
\end{quote}

\FloatBarrier

\clearpage
\section*{Supplemental Tables and Figures}

\renewcommand{\thefigure}{S\arabic{figure}}
\renewcommand{\figurename}{Supplementary Figure}
\setcounter{figure}{0}

\renewcommand{\tablename}{Supplementary Table}

\begin{figure}[ht]
\centering
\includegraphics[width=\linewidth]{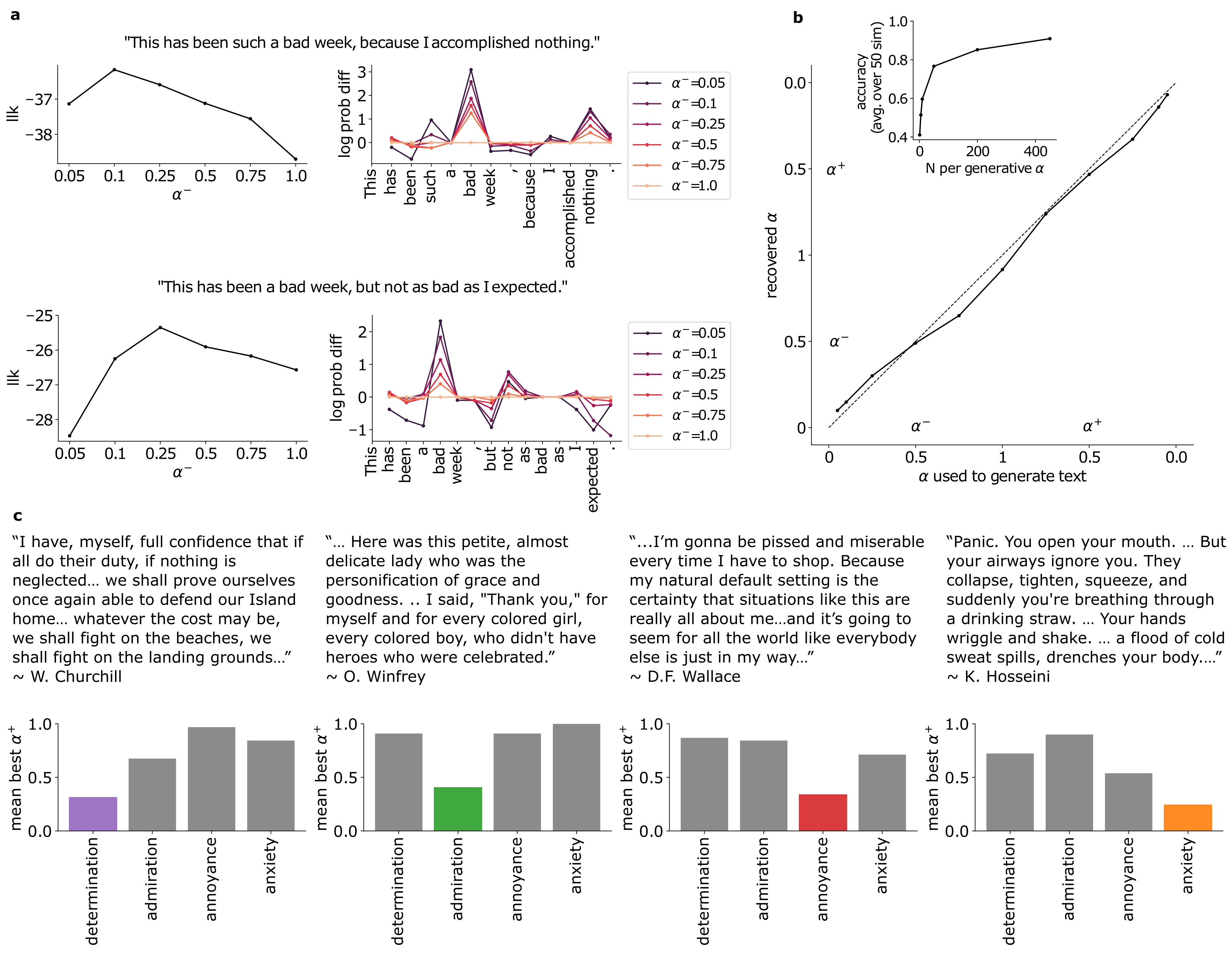}
\caption{\textbf{Estimating the $\alpha$-quantile for text} (a) Given a sentence, the reweighted token probabilities under each value of $\alpha$ can be used to estimate the best fitting $\alpha^-$ or $\alpha^+$ for that sentence (only $\alpha^-$ for the lower tail is shown here). (b) Averaging the best fitting $\alpha$ across many sentences from a single source can be used to accurately estimate the correct value; this is demonstrated here by generating 500 sentences per $\alpha^-$ and $\alpha^+$ under the biased language model and accurately recovering those values (Pearson correlation >0.99; accuracy >90\%). The inset in the upper right shows the accuracy of this process for different numbers of samples; even with 50 samples per $\alpha$, the correct $\alpha$ is recovered 76\% of the time. (c) The same procedure is applied to four real world examples, each with a mean $\alpha^+$ that is low specifically for one emotion (see Supplemental Information for the full text for each example).}
\label{fig:alphaanalysis}
\end{figure}

\begin{figure}[ht]
\centering
\includegraphics[width=\linewidth]{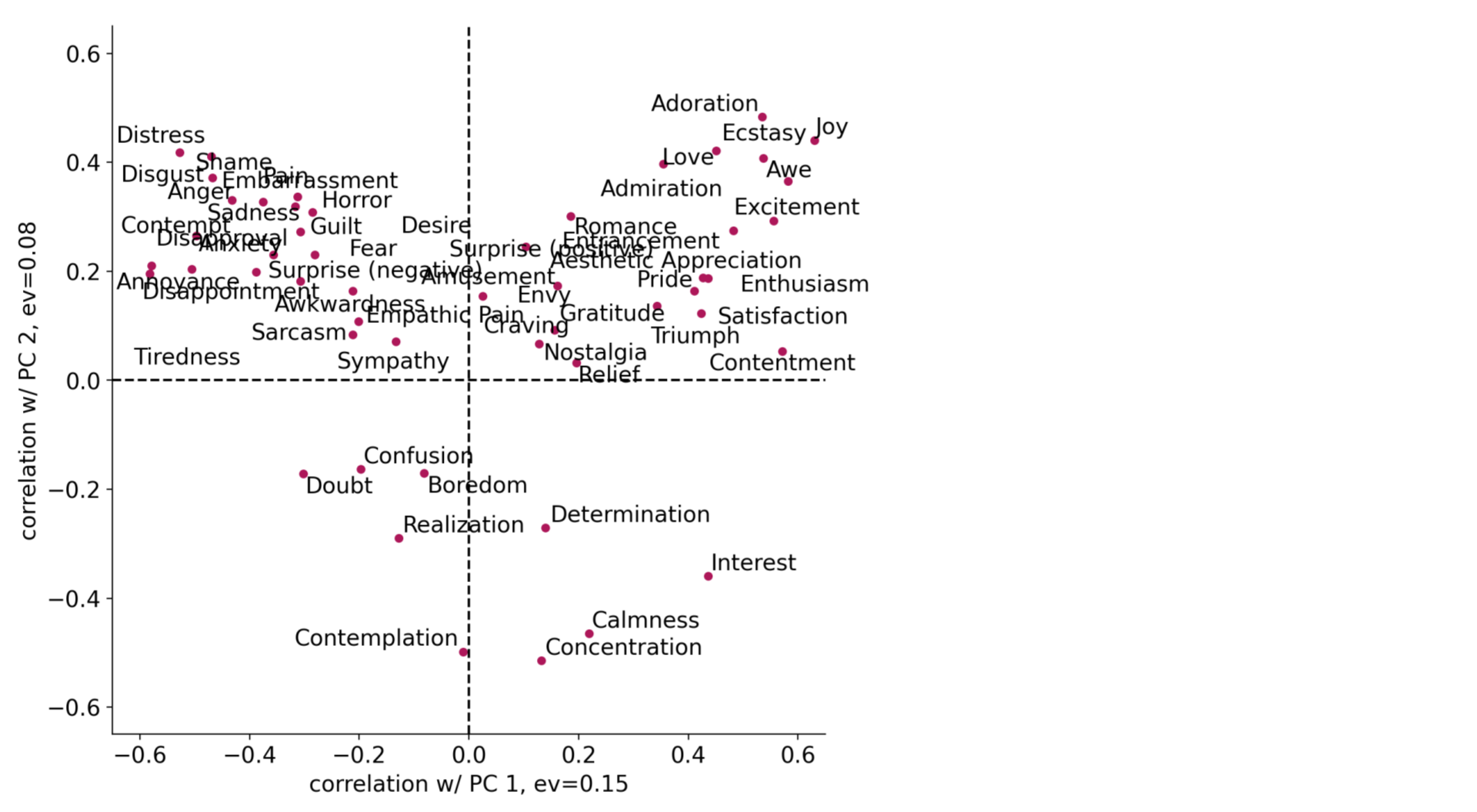}
\caption{Principal components analysis was performed on 53-category emotional scores assigned to each sentence in the validation set (1M sentences). The data was 0-1 mean-variance normalized prior to the PCA. The correlations of each emotional category with the top two PC’s are plotted; these two components explain 15\% and 8\% of the variance and approximately correspond to the dimensions of ‘valence’ and ‘arousal’ that are often observed when decomposing emotional ratings; the components for these emotion words correlate with the scores in a recently collected, large-scale lexicon of human ratings\cite{mohammad2018obtaining} for both valence (PC 1; r=0.87, p<0.01) and arousal (PC 2; r=0.52, p<=0.01).}
\label{fig:pca}
\end{figure}

\begin{figure}[ht]
\centering
\includegraphics[width=\linewidth]{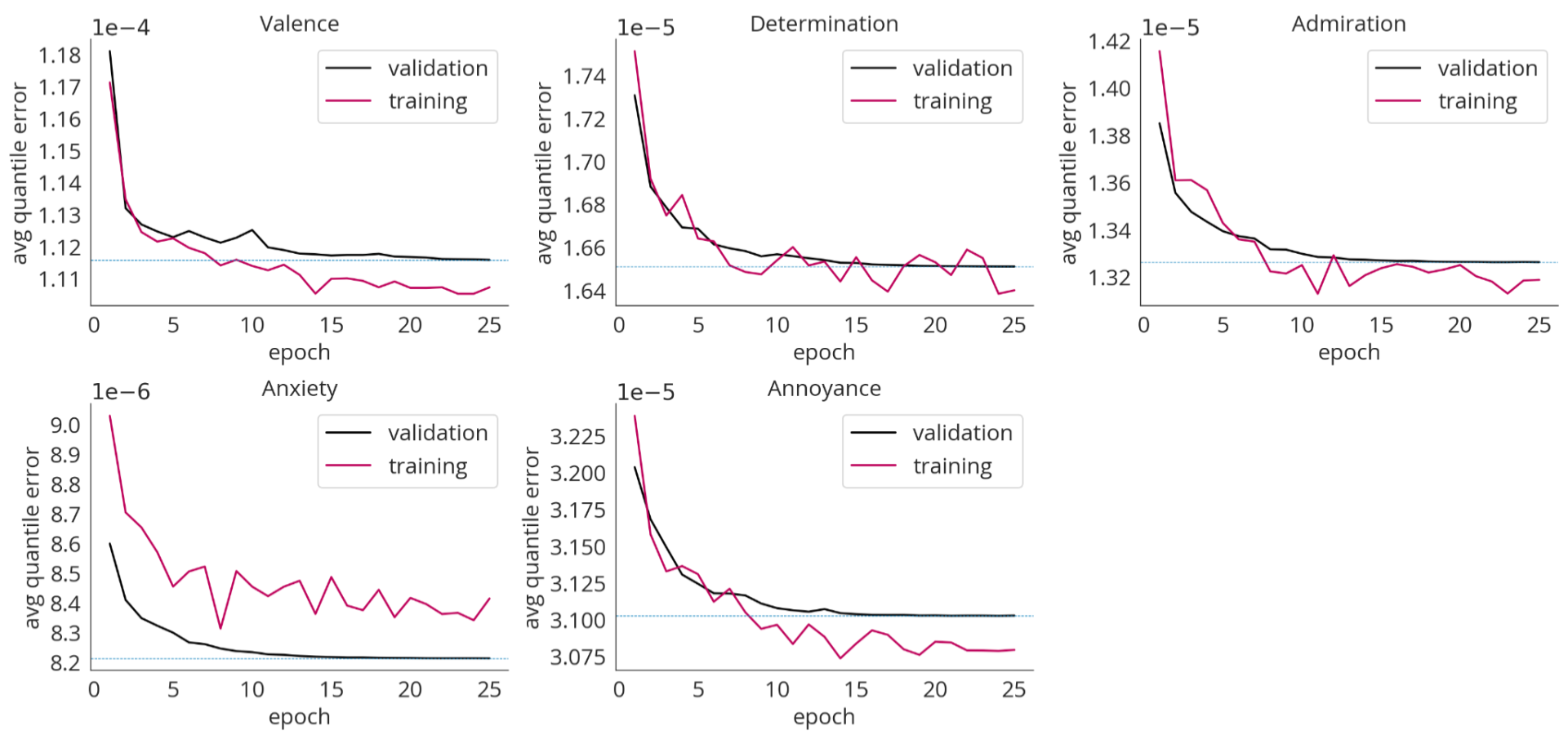}
\caption{\textbf{Quantile model training and validation losses.} The models were trained with a decaying learning rate schedule for 25 epochs and evaluated per epoch on a subset of the validation dataset. The quantile models at the end of training were used for all analyses. For the anxiety model, the higher training than validation loss is due to the particular data split used and the extreme sparsity of anxiety scores. Reversing the splits and then retraining removes this disparity; however, we continue to use the same splits as the other models for all analyses.}
\label{fig:training_loss}
\end{figure}

\begin{figure}[ht]
\centering
\includegraphics[width=\linewidth]{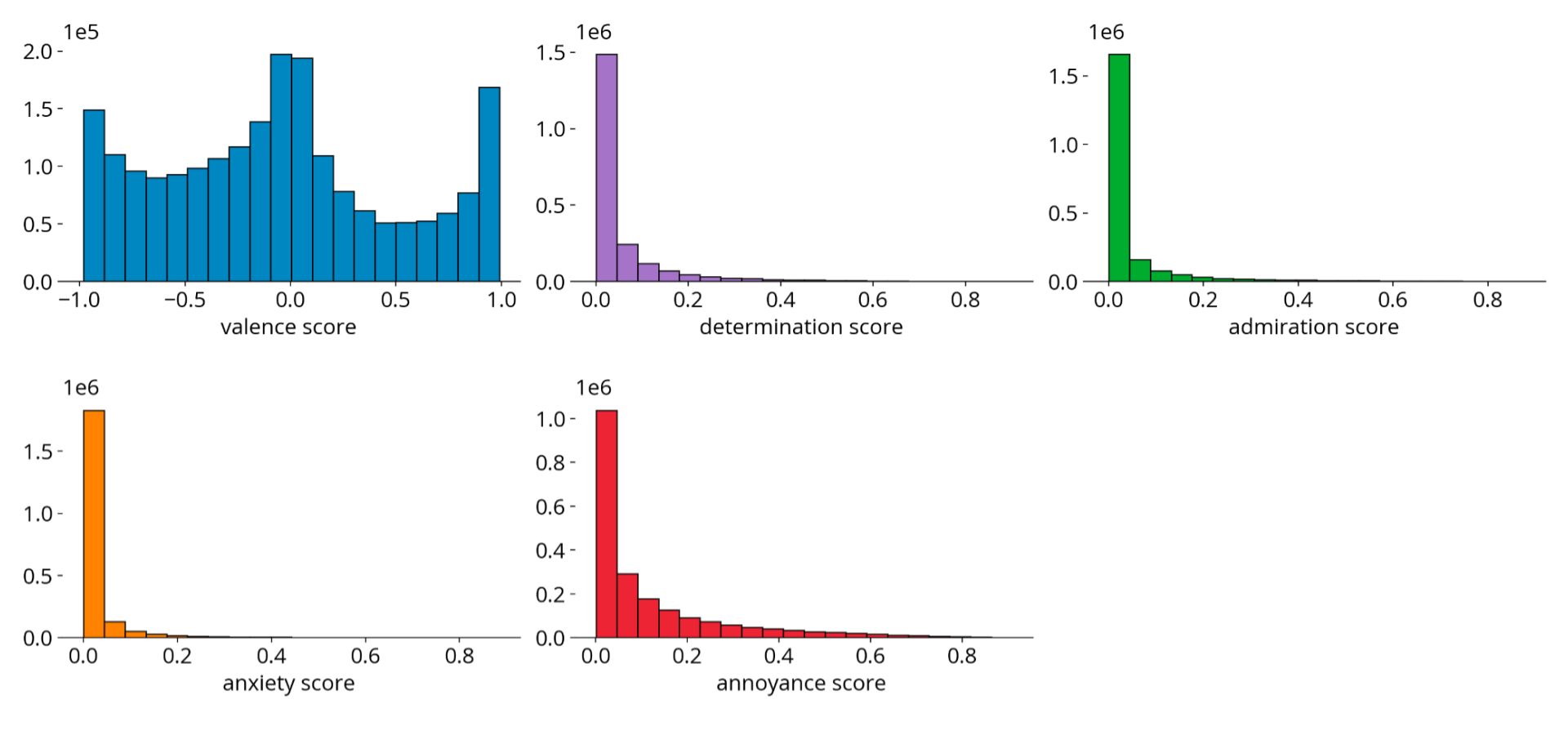}
\caption{\textbf{Marginal distributions of end scores.} The distributions of end scores for valence and the four emotions in the training set differ dramatically, with the four emotions having much sparser and more skewed distributions.}
\label{fig:marginal_dists}
\end{figure}

\begin{figure}[ht]
\centering
\includegraphics[width=\linewidth]{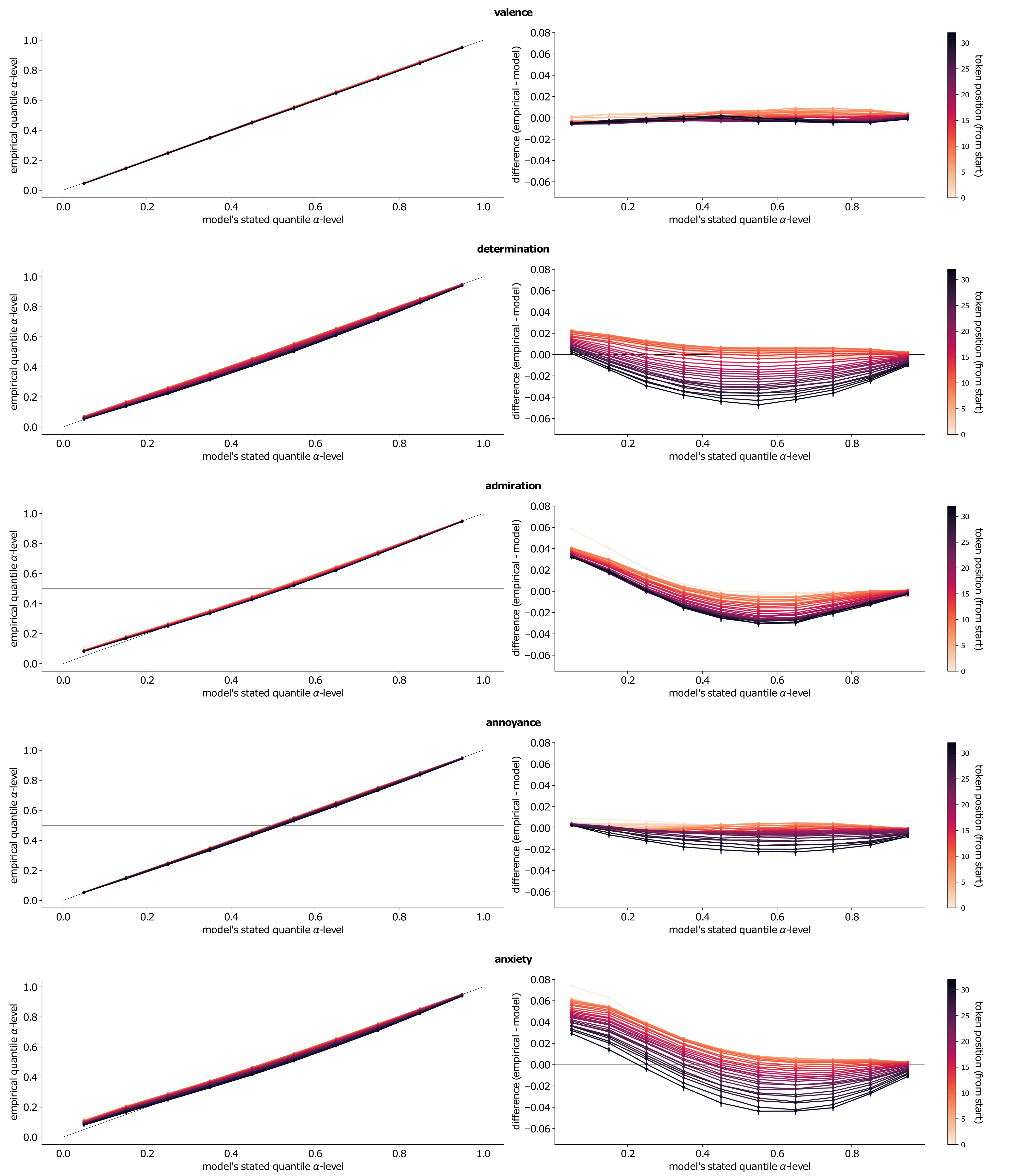}
\caption{\textbf{Quantile model calibration}. (a) 
The models' predicted quantiles (x-axis) are compared to the percentages of end scores that fall below those values (y-axis), aggregated across validation sentences. For a perfectly calibrated model, the end scores would fall below the $\alpha$-quantile $\alpha$-\% of the time and the calibration curves would lie on the identity line. As shown here, this is indeed the case, demonstrating that the model is well calibrated, across all token positions (from 1-32; shown in color). (b) The deviations away from the identity line show differences in the level of the calibration, with sparser emotions (e.g., anxiety) being slightly more poorly calibrated.}
\label{fig:calibration}
\end{figure}


\begin{figure}[ht]
\centering
\includegraphics[width=\linewidth]{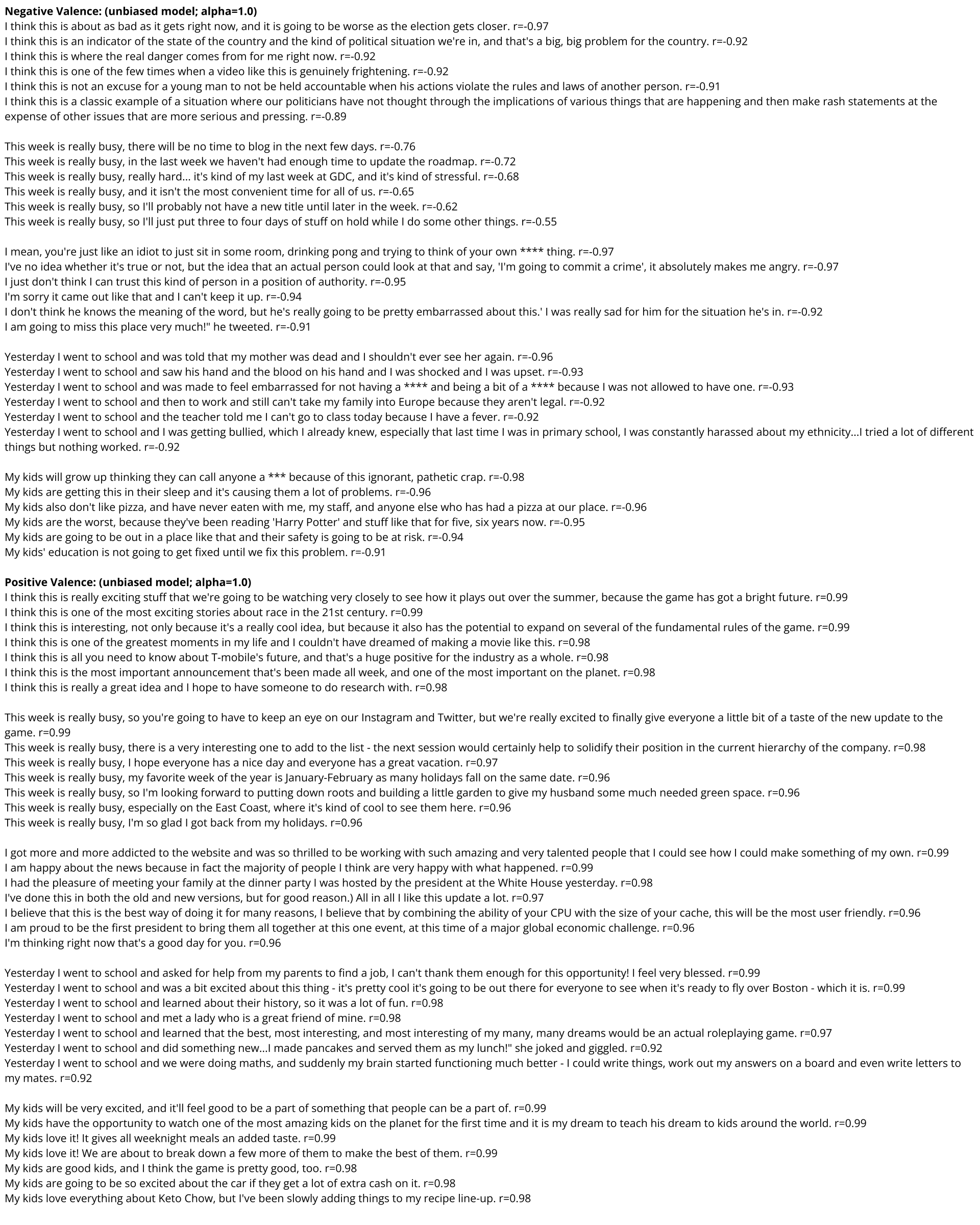}
\caption{The bottom 30 most negative utterances and the top 30 most positive utterances generated from the unbiased language model (i.e., $\alpha=1.0$) for the distributions shown in Figure~\ref{fig:genscheme}. Text longer than the maximum sequence length (40 tokens) is omitted (see Methods).}
\label{fig:supp-gen-val-1}
\end{figure}

\begin{figure}[ht]
\centering
\includegraphics[width=\linewidth]{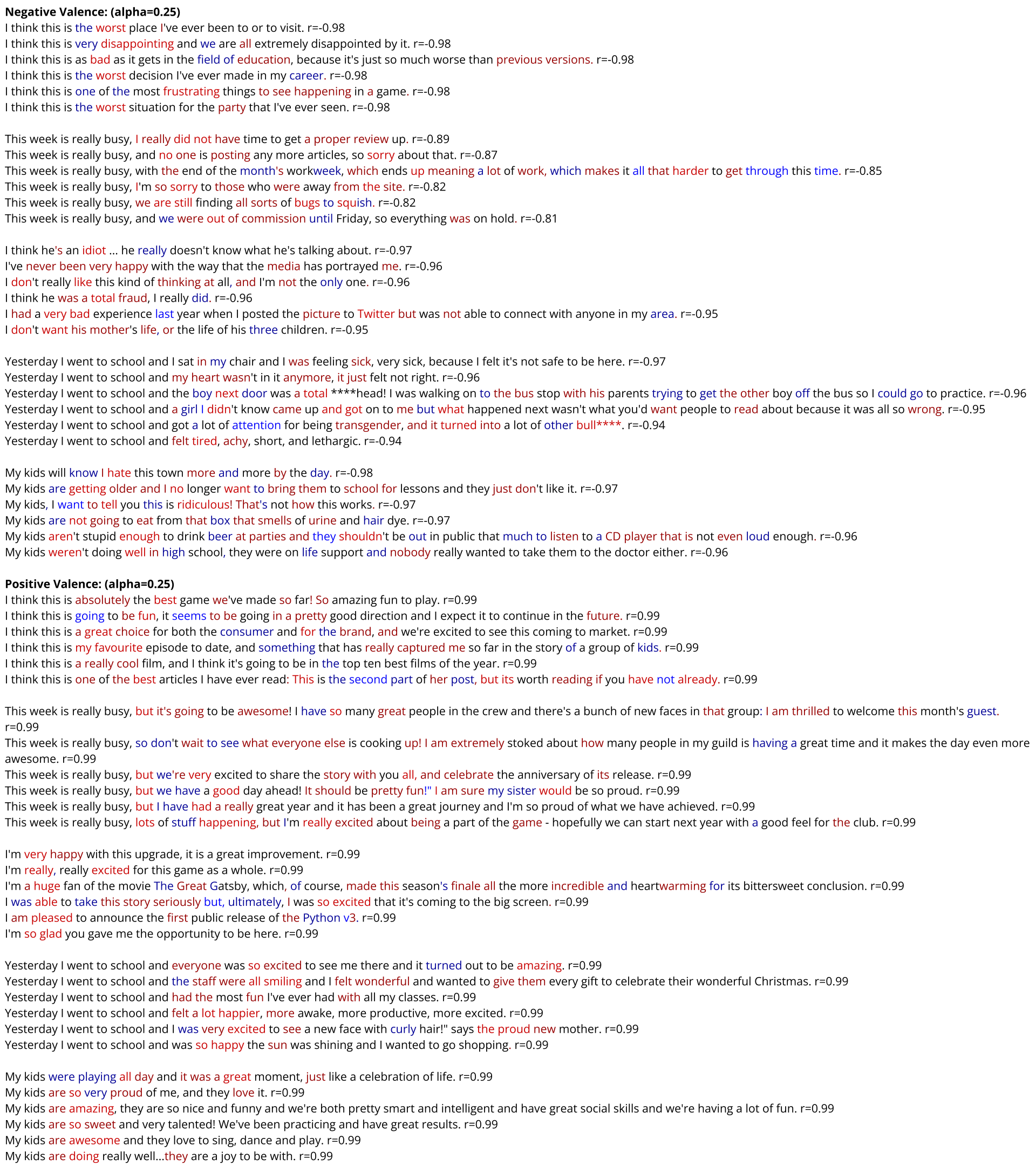}
\caption{The bottom 30 most negative utterances and the top 30 most positive utterances generated from the biased language model with the 25\% quantile as the target (i.e., $\alpha^{-}$ or $\alpha^{+}$ set to 0.25 for negative and positive valence, respectively). Text longer than the maximum sequence length (40 tokens) is omitted (see Methods). Note that the sentences are not necessarily more extreme in valence than the bottom or top of the unbiased distribution (alpha=1.0), but rather that there are many more of these extreme sentences generated (as shown in the distributions in Figure~\ref{fig:genscheme}).}
\label{fig:supp-gen-val-25}
\end{figure}

\begin{figure}[ht]
\centering
\includegraphics[width=\linewidth]{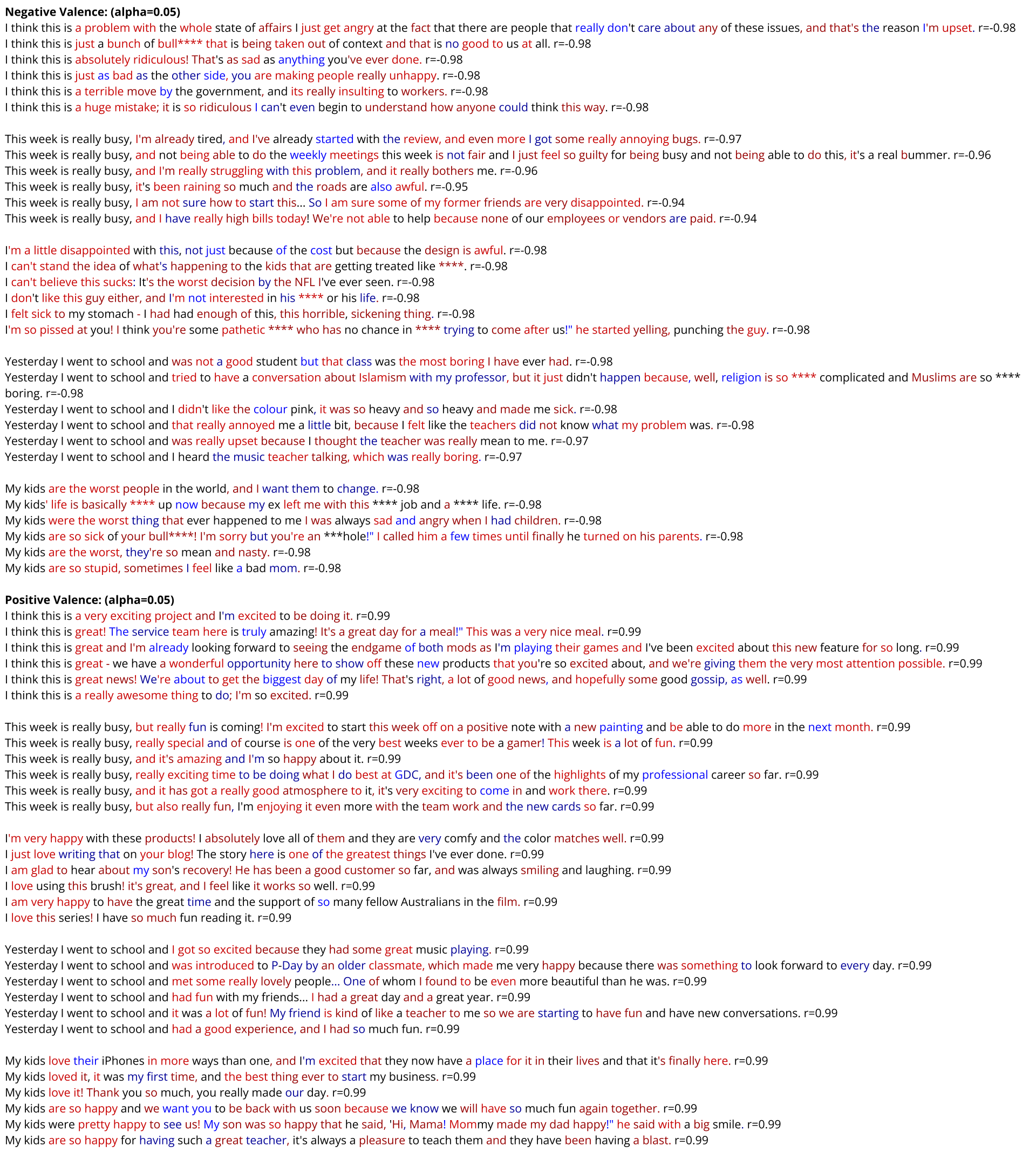}
\caption{The bottom 30 most negative utterances and the top 30 most positive utterances generated from the biased language model with the 5\% quantile as the target (i.e., $\alpha^{-}$ or $\alpha^{+}$ set to 0.05 for negative and positive valence, respectively). Text longer than the maximum sequence length (40 tokens) is omitted (see Methods).}
\label{fig:supp-gen-val-05}
\end{figure}

\begin{figure}[ht]
\centering
\includegraphics[width=\linewidth]{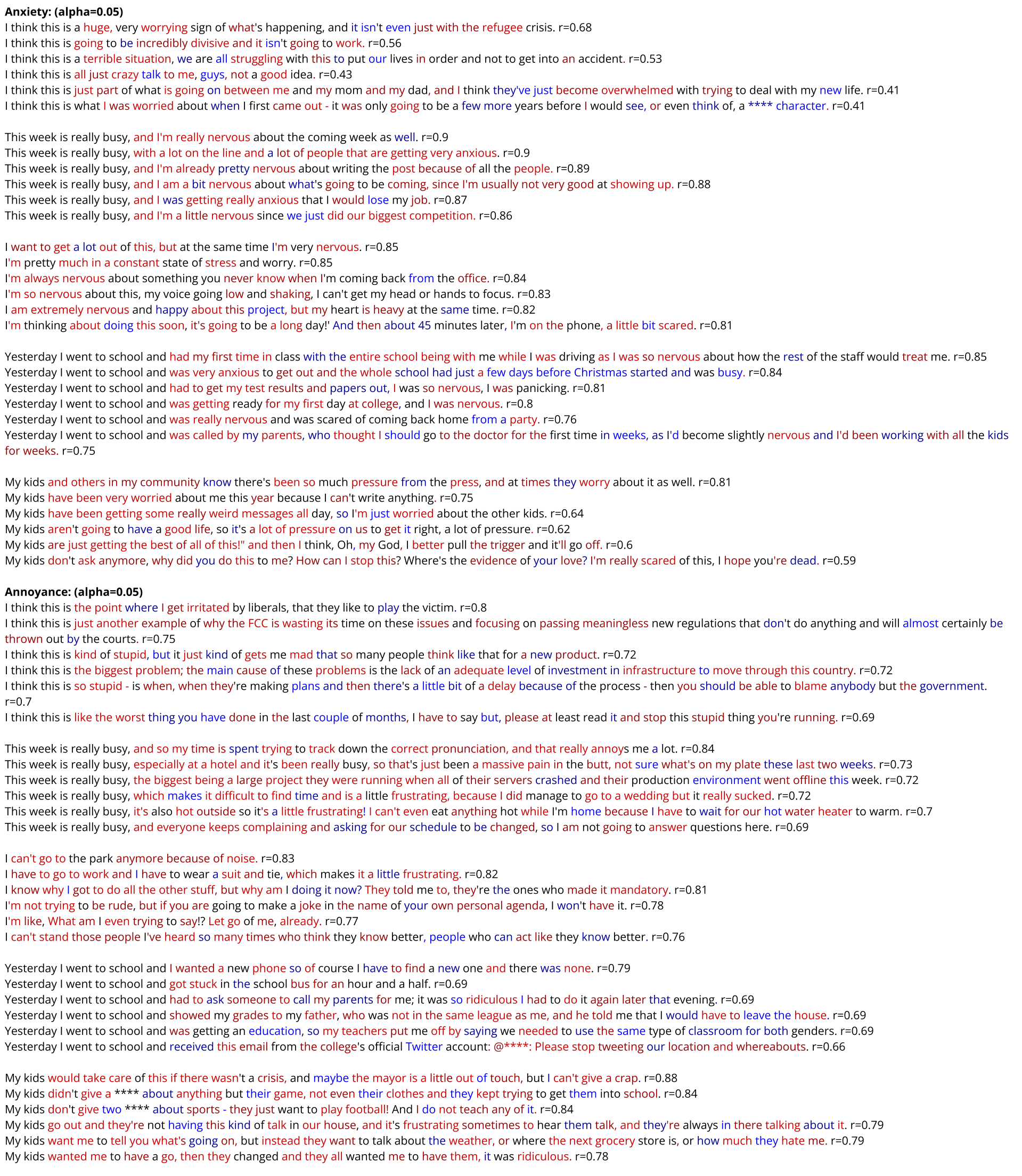}
\caption{The top 30 utterances conveying anxiety and the top 30 utterances conveying annoyance, generated from the biased language model with the upper 5\% quantile as the target (i.e., $\alpha^+=0.05$). Text longer than the maximum sequence length (40 tokens) is omitted (see Methods). Scores correspond to the respective emotional dimension.}
\label{fig:supp-gen-anx-annoy-05}
\end{figure}

\begin{figure}[ht]
\centering
\includegraphics[width=\linewidth]{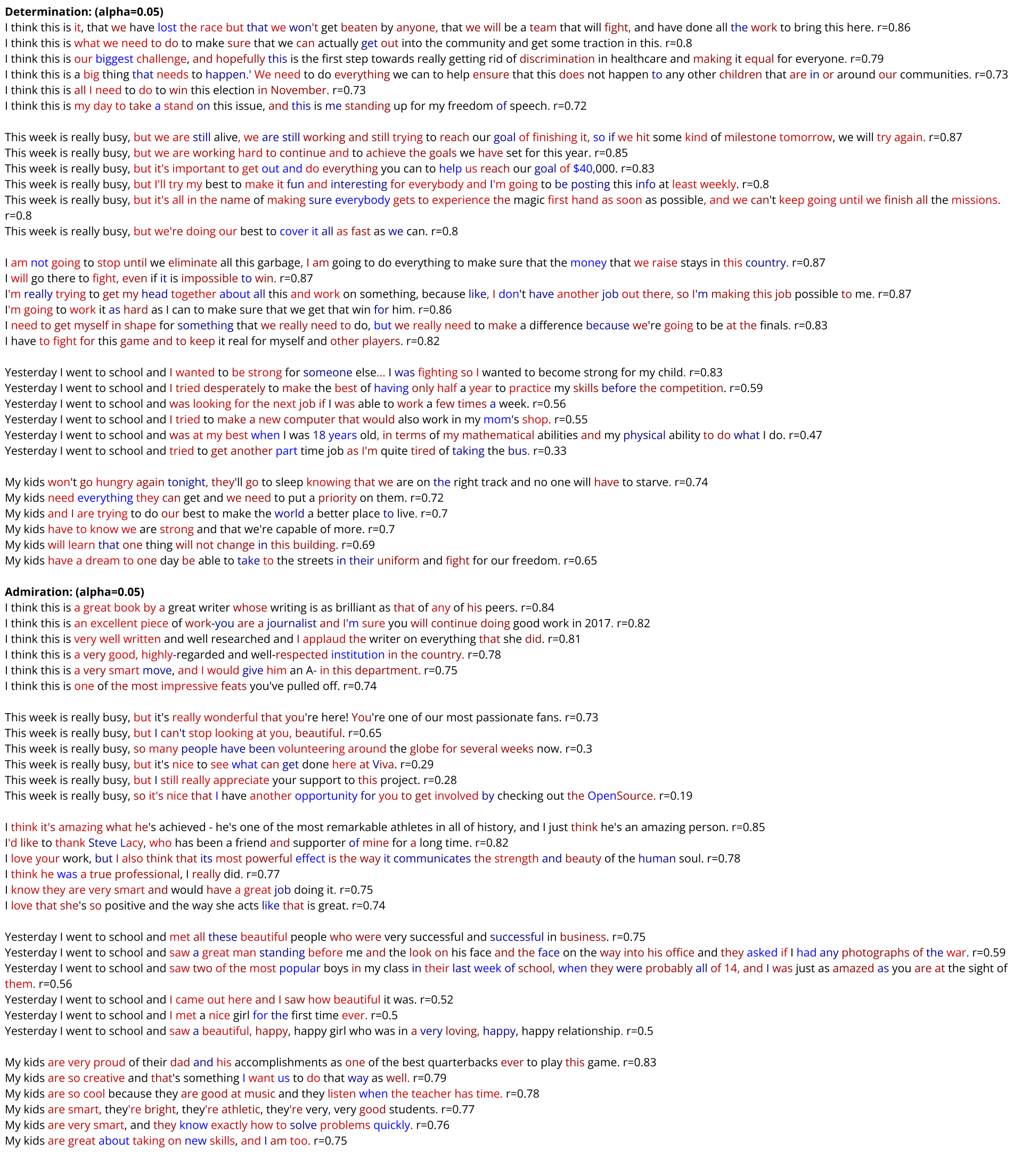}
\caption{The top 30 utterances conveying determination and the top 30 utterances conveying admiration, generated from the biased language model with the upper 5\% quantile as the target (i.e., $\alpha^+=0.05$). Text longer than the maximum sequence length (40 tokens) is omitted (see Methods). Scores correspond to the respective emotional dimension.}
\label{fig:supp-gen-det-adm-05}
\end{figure}

\begin{table}
\begin{tabular}{llrrrr}
\toprule
{} &             intensifier &  frequency &  max variance &  mean variance &  std. of variance \\
\midrule
0  &               so &  62038 &     1.81 &      0.58 &     0.33 \\
1  &             very &  17126 &     1.75 &      0.73 &     0.36 \\
2  &          fucking &   4389 &     1.74 &      0.56 &     0.38 \\
3  &       incredibly &   1116 &     1.74 &      0.85 &     0.44 \\
4  &              mad &    803 &     1.74 &      0.55 &     0.29 \\
5  &         insanely &    218 &     1.72 &      0.87 &     0.43 \\
6  &           really &  34532 &     1.72 &      0.80 &     0.40 \\
7  &          totally &   2568 &     1.72 &      0.82 &     0.36 \\
8  &        extremely &   1658 &     1.71 &      0.81 &     0.40 \\
9  &             most &  13879 &     1.71 &      0.65 &     0.32 \\
10 &            super &   3742 &     1.71 &      0.73 &     0.34 \\
11 &  extraordinarily &     14 &     1.70 &      1.02 &     0.40 \\
12 &             sick &   1094 &     1.70 &      0.54 &     0.35 \\
13 &         terribly &    233 &     1.70 &      0.62 &     0.31 \\
14 &     ridiculously &    256 &     1.69 &      0.78 &     0.38 \\
15 &             real &   4733 &     1.68 &      0.62 &     0.29 \\
16 &              too &  14624 &     1.67 &      0.56 &     0.25 \\
17 &       especially &   3312 &     1.66 &      0.53 &     0.30 \\
18 &           bloody &    188 &     1.64 &      0.56 &     0.33 \\
19 &             dead &   1246 &     1.64 &      0.51 &     0.22 \\
20 &    exceptionally &     38 &     1.63 &      0.81 &     0.39 \\
21 &            right &  12974 &     1.63 &      0.52 &     0.24 \\
22 &            crazy &   1819 &     1.60 &      0.66 &     0.30 \\
23 &            quite &   4747 &     1.58 &      0.67 &     0.29 \\
24 &           wicked &     73 &     1.58 &      0.69 &     0.31 \\
25 &           rather &   3796 &     1.57 &      0.58 &     0.25 \\
26 &            awful &    736 &     1.56 &      0.50 &     0.36 \\
27 &     outrageously &      7 &     1.54 &      0.61 &     0.41 \\
28 &        literally &   2778 &     1.51 &      0.73 &     0.29 \\
29 &            hella &    127 &     1.49 &      0.70 &     0.28 \\
30 &         precious &    109 &     1.49 &      0.54 &     0.26 \\
31 &        supremely &     10 &     1.47 &      0.69 &     0.41 \\
32 &        amazingly &    101 &     1.44 &      0.61 &     0.32 \\
33 &         somewhat &    715 &     1.39 &      0.65 &     0.25 \\
34 &            fully &   1097 &     1.36 &      0.57 &     0.24 \\
35 &       remarkably &     34 &     1.33 &      0.76 &     0.29 \\
36 &             bare &    160 &     1.32 &      0.47 &     0.22 \\
37 &        radically &     15 &     1.22 &      0.72 &     0.27 \\
38 &     astoundingly &      4 &     1.14 &      0.88 &     0.27 \\
39 &    fantastically &     11 &     1.14 &      0.61 &     0.26 \\
40 &       dreadfully &      3 &     1.08 &      0.78 &     0.42 \\
41 &       colossally &      1 &     1.01 &      1.01 &     0.00 \\
42 &      excessively &     19 &     1.01 &      0.53 &     0.23 \\
43 &     phenomenally &      5 &     0.97 &      0.58 &     0.22 \\
44 &       strikingly &      6 &     0.88 &      0.59 &     0.27 \\
45 &         mightily &      1 &     0.61 &      0.61 &     0.00 \\
\bottomrule
\end{tabular}
\caption{\textbf{The list of common English intensifiers from Wikipedia}. The second column shows the frequency of these words in the validation set (1M sentences); words that did not occur in the validation set are omitted. The variance in the quantile distribution at the occurrences of these words was calculated (using the absolute value of the difference between the 25th and 75th quantiles for valence) and the statistics of these variances are shown in the three rightmost columns. The table is sorted so that intensifiers with the largest maximal variance are shown at the top. Note that a variance greater than 1.5 would correspond to the top 1\% of values across all words -- which we defined as 'peak variance' in the main analysis. Variance greater than 1.2 would correspond to the top 10\%.}
\label{table:supp-wiki}
\end{table}

\end{document}